\documentclass[10pt,twocolumn,letterpaper]{article}

\usepackage{arxiv}
\usepackage{times}
\usepackage{epsfig}
\usepackage{graphicx}
\usepackage{amsmath}
\usepackage{amssymb}
\usepackage{comment}
\usepackage[toc,page]{appendix}
\newcommand{\mb}[1] {\mathbf{#1}}
\newcommand{\cframe} {canonical frames}
\newcommand{\cframes} {canonical frames}

\newcommand{\ccff} {Canonical Frames}
\newcommand{\jw}[1]{#1}

\usepackage[pagebackref=true,breaklinks=true,letterpaper=true,colorlinks,bookmarks=false]{hyperref}

\arxivfinalcopy 

\def\arxivPaperID{****} 

\begin{document}
\title{FrameNet: Learning Local \ccff{} of 3D Surfaces\\
from a Single RGB Image}

\author{
Jingwei Huang$^{1}$ \qquad Yichao Zhou$^{2}$  \qquad Thomas Funkhouser$^{3,4}$ \qquad Leonidas Guibas$^{1}$
\vspace{0.2cm} \\ 
$^{1}$Stanford University \qquad $^{2}$University of California, Berkeley \qquad $^{3}$Princeton University \quad $^{4}$Google \vspace{-0.3cm}
}

\maketitle

\begin{abstract}
   In this work, we introduce the novel problem of identifying dense canonical 3D coordinate frames from a single RGB image. We observe that each pixel in an image corresponds to a surface in the underlying 3D geometry, where a canonical frame can be identified as represented by three orthogonal axes, one along its normal direction and two in its tangent plane.  We propose an algorithm to predict these axes from RGB.   Our first insight is that \cframe{} computed automatically with recently introduced direction field synthesis methods can provide training data for the task.  Our second insight is that networks designed for surface normal prediction provide better results when trained jointly to predict \cframe{}, and even better when trained to also predict 2D projections of \cframe{}.   We conjecture this is because projections of canonical tangent directions often align with local gradients in images, and because those directions are tightly linked to 3D \cframe{} through projective geometry and orthogonality constraints.   In our experiments, we find that our method predicts 3D \cframe{} that can be used in applications ranging from surface normal estimation, feature matching, and augmented reality.


\end{abstract}

\section{Introduction}

In recent years, learning to predict 3D properties from a single RGB image has made great progress.  For example, monocular depth estimation~\cite{shelhamer2015scene,li2017two,xu2017multi,wang2018adaptive,fu2018deep} and surface normal prediction~\cite{eigen2015predicting,wang2015designing,bansal2016marr,qi2018geonet} have improved dramatically.  There are many applications of them in scene understanding and robot interaction.

The main challenge in this domain is choosing an appropriate representation of 3D geometry to predict.  Zhang~\textit{et al.}~\cite{zhang2018deep} predicts dense surface normals and then uses geometric constraints to solve for depths from them with a global optimization.  GeoNet~\cite{qi2018geonet} predicts both surface normals and depths and then passes them to a refinement network for further optimization.  These methods are clever in their use of geometric constraints to regularize dense predictions.   However, still they infer only 2 of the 3 degrees of freedom in a 3D coordinate frame -- the rotation in the tangent plane around the surface normal is left unknown.  As such, they are missing 3D information critical to many applications.   For example, they cannot assist an AR system in placing a picture frame on a wall or a laptop on a table because they don't know the full 3D coordinate frame (including tangent directions) of the wall and table surfaces.

 
 \begin{figure}
    \centering
    \includegraphics[width=\linewidth,height=0.75\linewidth]{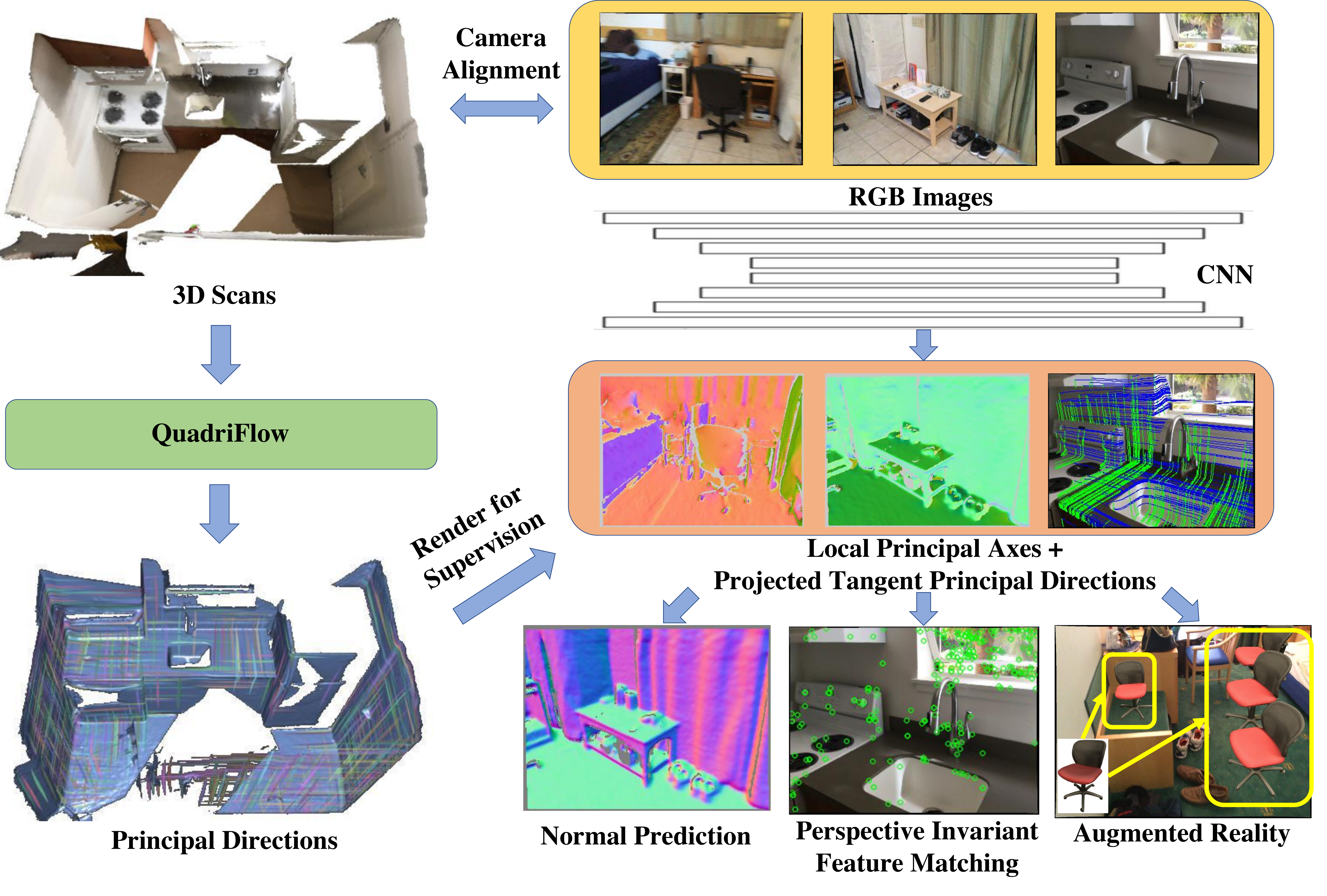}
    \caption{We propose the novel task of learning local \cframe{} from a single image. We compute the \cframe{} from meshes and render them to images to supervise the learning. The joint estimation lead to better surface normal estimation. The tangent principal axes can be used for perspective-invariant feature descriptors and enable new applications for augmented reality.}
    \label{fig:teaser}
\end{figure}

\begin{figure}
    \centering
     \begin{minipage}{0.49\linewidth}
     \centering
     \includegraphics[width=\linewidth,height=0.75\linewidth]{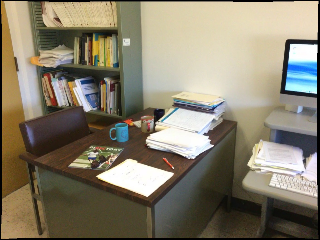}\\
     (a) RGB
     \end{minipage}
     \begin{minipage}{0.49\linewidth}
     \centering
     \includegraphics[width=\linewidth,height=0.75\linewidth]{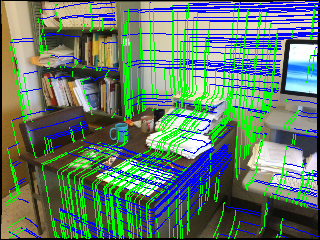}\\
     (b) Principal Directions
     \end{minipage}
    \caption{We visualize the principal directions by picking random seed points and tracing along two projected principal directions in the image plane. The projected principal directions usually follow the texture directions or object boundaries.}
    \label{fig:vis-direction}
\end{figure}

In this work, we propose a novel image-to-3D task: pixel-wise local \cframe{} estimation from a single image (figure~\ref{fig:teaser}). The orthogonal bases of the \cframe{} include the surface normal plus two tangent principal directions generally following the principal curvatures of the surface. With an understanding of the local \cframe{}, our task not only offers better surface normal estimation but also enables perspective-invariant feature descriptors and new applications in augmented reality. \jw{We propose to jointly estimate the projected tangent directions and \cframe{}, and find that the joint estimation can improve the estimation of tangent principal directions and surface normals. We believe the reason is that tangent principal directions can be directly inferred from low-level features such as image gradients, since their projections are often aligned with the texture directions or object boundaries, as shown in figure~\ref{fig:vis-direction}(b).}

 
In addition to surface normal estimation, the tangent principal directions are also important 3D properties, as they imply the canonical transformation that maps the 3D surface to the image plane. Thus, an inverse transform of the image patch could alleviate the perspective distortions. We show that this can improve local patch descriptors like SIFT~\cite{lowe2004distinctive} which is rotational and scale invariant but influenced by perspective. Further, understanding the perspective enables new applications in augmented reality, including inserting new elements on top of either flat or curved surfaces in the scene with the correct perspective distortion.

To make the learning practical, we need the ground truth local \cframe{} labeling for RGB frames. We propose to use existing 3D reconstruction datasets like ScanNet~\cite{dai2017scannet} which offers alignment between images and 3D meshes. We can compute the \cframe{} on the meshes and render them to the aligned RGB frames. \jw{We follow TextureNet~\cite{huang2018texturenet} that computes tangent principal directions as the extrinsic four-way rotationally symmetric (\textit{(4-RoSy) orientation field}) using QuadriFlow~\cite{huang2018quadriflow}.}  We find that the surface tangent directions computed this way are consistent enough to be learned by a network.

Overall, the core contributions of the paper are:
    \vspace{-0.1in}
\begin{itemize}
    \item Identifying an important new 3D vision problem: local \cframe{} estimation from RGB images.
    \vspace{-0.1in}
    \item Using projected tangent principal directions to improve \cframe{} estimation, outperforming existing works on surface normal estimation.
    \vspace{-0.1in}
    \item Exploiting tangent projected principal directions to compute perspective invariant feature descriptors.
    \vspace{-0.1in}
    \item Inserting new elements in the scene in a manner aware of perspective distortions, for augmented reality.
\end{itemize}

\section{Related Work}
\paragraph{3D from Single Image.}
Estimating 2.5D geometry properties from a single image has become popular in recent years. Traditional methods aim at understanding low-level image information and geometry constraints. For example, Torralba \textit{et al.}~\cite{torralba2002depth} exploits the scene structure to estimate the absolute depth values. Saxena \textit{et al.}~\cite{saxena2006learning} uses hand-crafted features to predict the depth based on Markov random fields. Hoiem \textit{et al.}~\cite{hoiem2007recovering} recovers scene layout guided by the vanishing points and lines. Shi \textit{et al.}~\cite{shi2015break} estimates the defocus blur and uses it to assist depth estimation.

With the availability of large-scale dataset and the success of deep learning, many methods have been proposed for depth or/and surface normal estimation. For depth estimation, Eigen \textit{et al.}~\cite{eigen2014depth} uses CNN to predict indoor depth maps on the NYUv2 dataset. With the powerful backbone network architecture like VGG~\cite{simonyan2014very} or ResNet~\cite{he2016deep}, depth estimation can be further improved~\cite{garg2016unsupervised,xie2016deep3d}. DORN~\cite{fu2018deep} proposes a novel ordinary loss and achieves the state-of-the-art in KITTI~\cite{geiger2013vision}. For surface normal estimation, Wang \textit{et al.}~\cite{wang2015designing} incorporate vanishing point and layout information in the network architecture. Eigen and Fergus~\cite{eigen2015predicting} trained a coarse-to-fine CNN to refine the details of the normals. The skip-connected architecture~\cite{bansal2016marr} is proposed to fuse hidden layers for surface normal estimation.

Since surface normal and depth are related to each other, another set of methods aimed at jointly predicting both to improve the performance. Wang \textit{et al.}~\cite{wang2016surge} exploits the consistency between normal and depth in planar regions. GeoNet~\cite{qi2018geonet} proposes a refinement network to enhance the depth and normal estimation from each other. Zhang \textit{et al.}~\cite{zhang2018deep} predict the normal and solve a global optimization problem to complete the depth.  We take a further step by jointly estimating all axes of a 3D canonical frame at each pixel, which helps both regularize the prediction through constraints and is useful in applications (see Sec. \ref{sec:applications}).

\paragraph{Local \ccff{}}
\jw{Computing local \cframe{} on surfaces is a fundamental step for many problems.} 3DLite~\cite{huang20173dlite} builds \cframe{} in fitted 3D planes for color optimizations. GCNN~\cite{masci2015geodesic} defines local frames with spherical coordinates and applies discrete patch operators on tangent planes. ACNN~\cite{boscaini2016learning} introduces the anisotropic heat kernels derived from principal curvatures \jw{so that it can apply convolutions in local canonical frames defined by principal axes. Such canonical frame} is also used in Xu \textit{et al.}~\cite{xu2017directionally} for nonrigid segmentation, by Tatarchenko \textit{et al.}~\cite{tatarchenko2018tangent,huang2018texturenet} for semantic segmentation of the 3D scenes. \jw{We aim at recognizing such canonical frames from 2D images, and we compute them from 3D surfaces to supervise the learning.}

TextureNet~\cite{huang2018texturenet} highlights the challenges of computing robust local \cframe{} at planar surface regions, where the principal curvatures are undetermined or highly influenced by noise or uneven sampling. Therefore, it proposes to compute a 4-RoSy orientation field to represent the principal directions. The 4-RoSy orientation field is an important concept in geometry processing community~\cite{ray2008n,lai2010metric}. The target directions are aligned with the principal curvatures~\cite{cohen2003restricted,cazals2005estimating}, but regularized by additional energy to vary smoothly. This can be achieved by optimizing a nonlinear energy by periodic functions~\cite{hertzmann2000illustrating,ray2009geometry} or a mixed-integer representation~\cite{ray2008n,bommes2009mixed}. In our work, we use QuadriFlow~\cite{huang2018quadriflow} to optimize the 4-RoSy field so that it aligns with the principal curvatures at the curved surface and ensures smoothness in flat regions (where principal directions are ill-defined) as well as robustness to noise.

\section{Approach}
In this section, we develop our approach for learning local \cframe{} from RGB images. First, we discuss the ground truth labeling of \cframe{} from 2D images in section~\ref{sec:prepare-data}. Then, we discuss the concept of projected tangent principal directions in section~\ref{sec:project}. Finally in section~\ref{sec:network}, we propose several energy terms that enforce the neural network to predict consistent local \cframe{} assisted by the projected tangent principal directions. Since we focus on the behavior of the local \cframe{} rather than the neural network architecture, we can adopt any neural network that predicts per-pixel features (see experiments in Sec. \ref{sec:evaluation} and \ref{sec:applications}).

\subsection{Local \ccff{} Generation}
\label{sec:prepare-data}
To label the \cframe, we need a dataset with 3D meshes aligned with RGB images so that we can compute frames from geometry and render them to images to produce ground truth. We choose ScanNet~\cite{dai2017scannet} for our experiments.

\begin{figure}
    \centering
    \includegraphics[width=\linewidth]{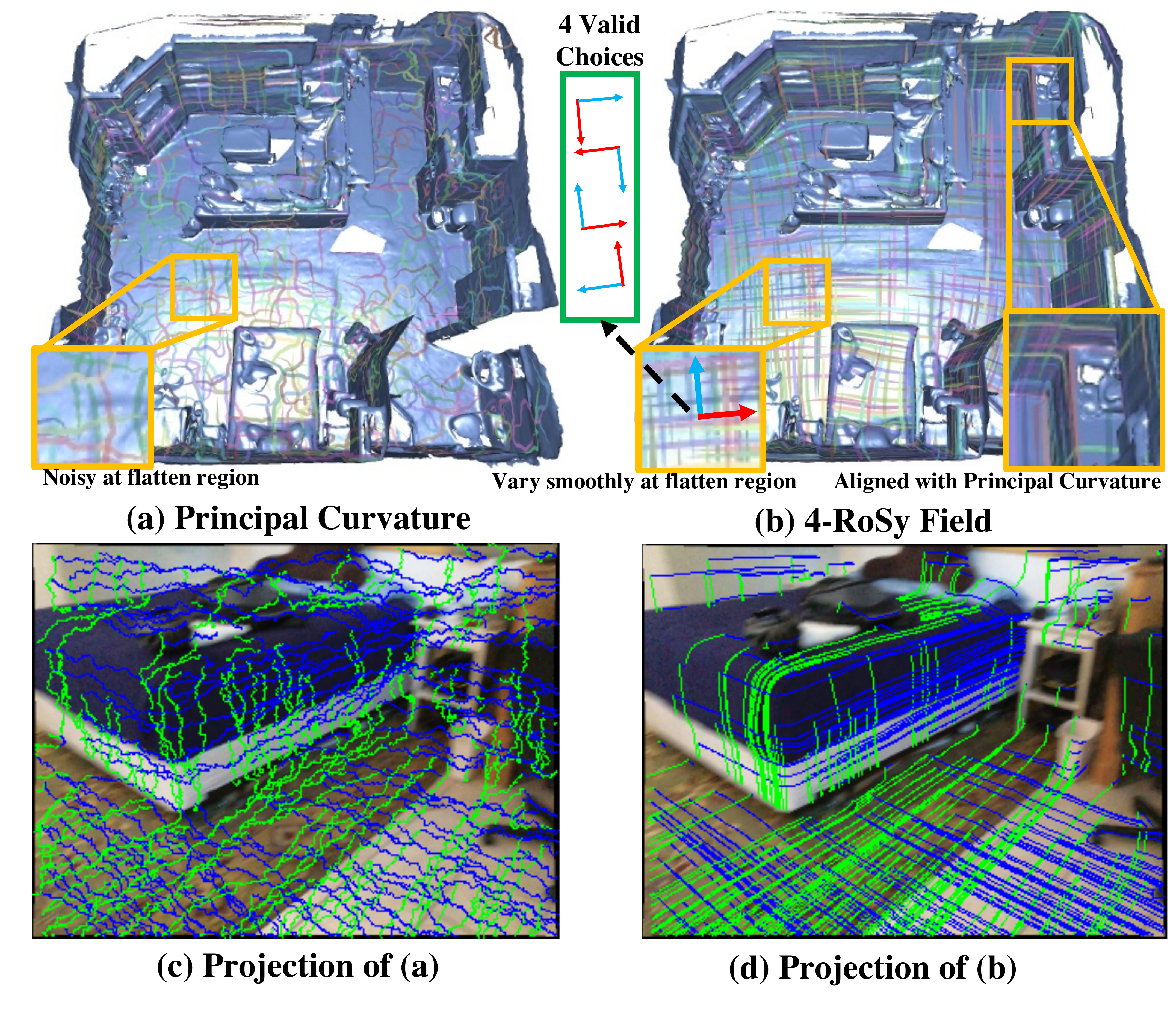}
    \vspace{-0.20in}
    \caption{(a) computes the direction field from estimated principal curvatures. Noises exist in both the geometry and the projections in images, as shown in (c). (b) computes the 4-RoSy field using QuadriFlow~\cite{huang2018quadriflow} and produces robust tangent principal directions, as shown in (d) as the projection in the image plane.}
    \label{fig:vis-geometry}
\vspace{-0.10in}
\end{figure}

We compute \cframe{} as surface normals and tangent principal directions with the scene geometry. It is straightforward to compute surface normals, but tangent principal directions at flat regions are hard to compute especially in the presence of noise. As visualized in figure~\ref{fig:vis-geometry}(a,c), the tangent principal directions can be pretty noisy. To solve this problem, we adopt the 4-RoSy field using QuadriFlow~\cite{huang2018quadriflow} as proposed by TextureNet~\cite{huang2018texturenet}, as shown in figure~\ref{fig:vis-geometry}(b,d): This field generates consistent directions which vary smoothly at flatter regions and are aligned with the principal curvatures at curved surfaces. The cross-field is 4-RoSy since there are four valid choices for the tangent principal directions at each vertex. Considering this, we pick any pair of orthogonal tangent vectors in the cross field to represent the principal directions, but we also view the other three alternatives as valid ground truth.

\begin{figure}
    \centering
     \begin{minipage}{0.19\linewidth}
     \centering
     \includegraphics[width=\linewidth]{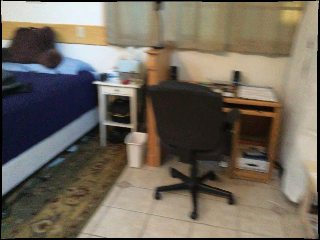}\\
     \includegraphics[width=\linewidth]{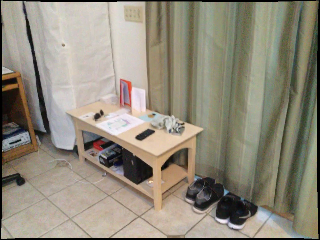}\\
     \includegraphics[width=\linewidth]{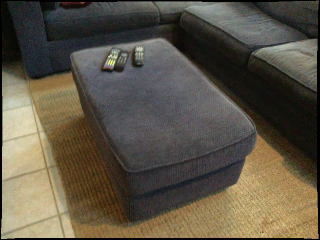}\\
     \includegraphics[width=\linewidth]{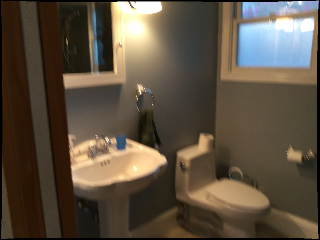}\\
     \vspace{-0.05in}
     RGB
    \end{minipage}
     \begin{minipage}{0.19\linewidth}
     \centering
     \includegraphics[width=\linewidth]{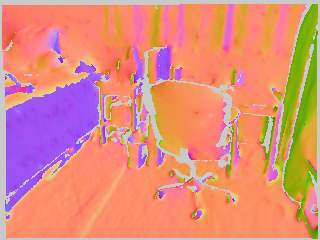}\\
     \includegraphics[width=\linewidth]{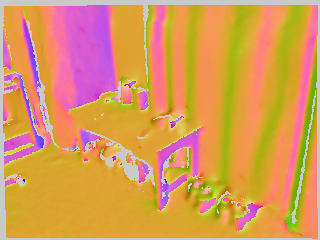}\\
     \includegraphics[width=\linewidth]{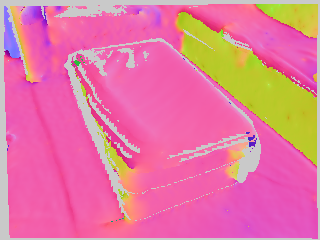}\\
     \includegraphics[width=\linewidth]{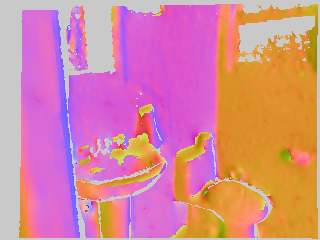}\\
     \vspace{-0.05in}
     X
    \end{minipage}
     \begin{minipage}{0.19\linewidth}
     \centering
     \includegraphics[width=\linewidth]{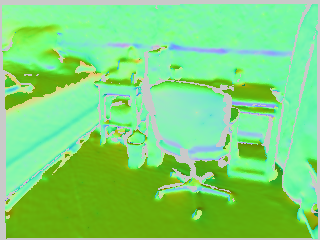}\\
     \includegraphics[width=\linewidth]{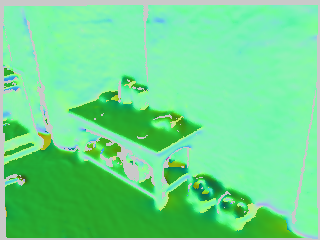}\\
     \includegraphics[width=\linewidth]{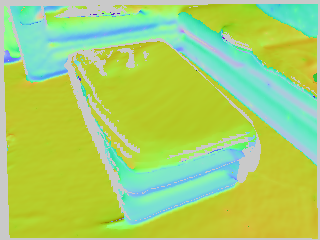}\\
     \includegraphics[width=\linewidth]{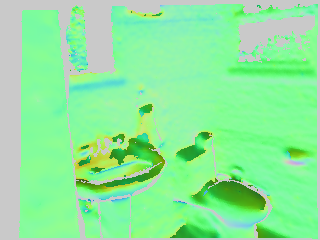}\\
     \vspace{-0.05in}
     Y
    \end{minipage}
     \begin{minipage}{0.19\linewidth}
     \centering
     \includegraphics[width=\linewidth]{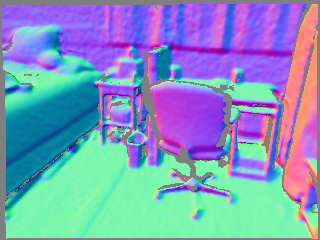}\\
     \includegraphics[width=\linewidth]{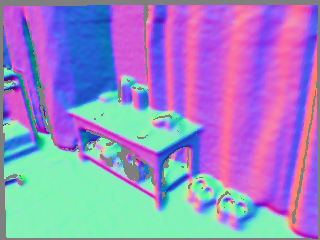}\\
     \includegraphics[width=\linewidth]{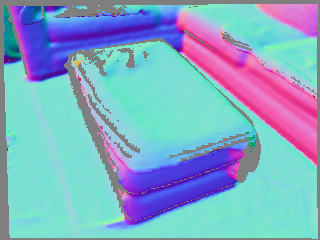}\\
     \includegraphics[width=\linewidth]{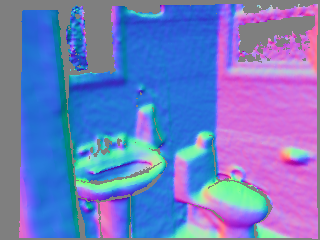}\\
     \vspace{-0.05in}
     Normal
    \end{minipage}
     \begin{minipage}{0.19\linewidth}
     \centering
     \includegraphics[width=\linewidth]{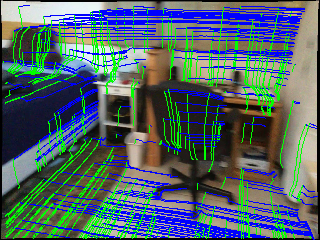}\\
     \includegraphics[width=\linewidth]{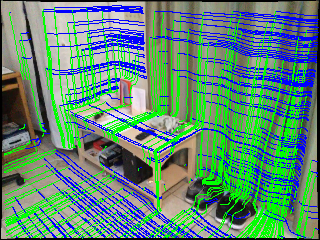}\\
     \includegraphics[width=\linewidth]{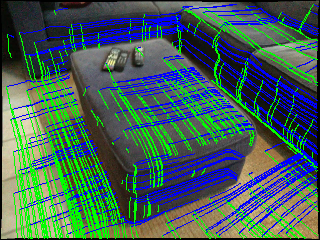}\\
     \includegraphics[width=\linewidth]{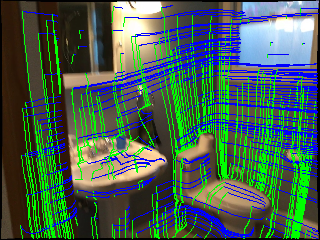}\\
     \vspace{-0.05in}
     Projection
    \end{minipage}
    \caption{Local \ccff{} Dataset. For each RGB frame, we render the corresponding tangent principal directions (X and Y) for each pixel. The surface normal can be computed as the cross product of the principal directions.}
    \label{fig:dataset}
\vspace{-0.2in}
\end{figure}

We store the computed local \cframe{} on top of mesh vertices and render them to images after transforming them to the camera space. For each triangle to be rendered, we enumerate the $90N(N\in \mathbb{Z})$ degree rotations to the tangent principal directions of the last two vertices, so as to align them with the first vertex before the standard rasterization stage. This is to deal with the 4-way rotational ambiguities in the cross field. For each RGB image, we render and save the tangent principal directions as two images, as shown in figure~\ref{fig:dataset} as X and Y. The ground truth normal can be directly computed as the cross product of them.

\begin{figure*}
    \centering
    \includegraphics[width=\linewidth,height=0.25\linewidth]{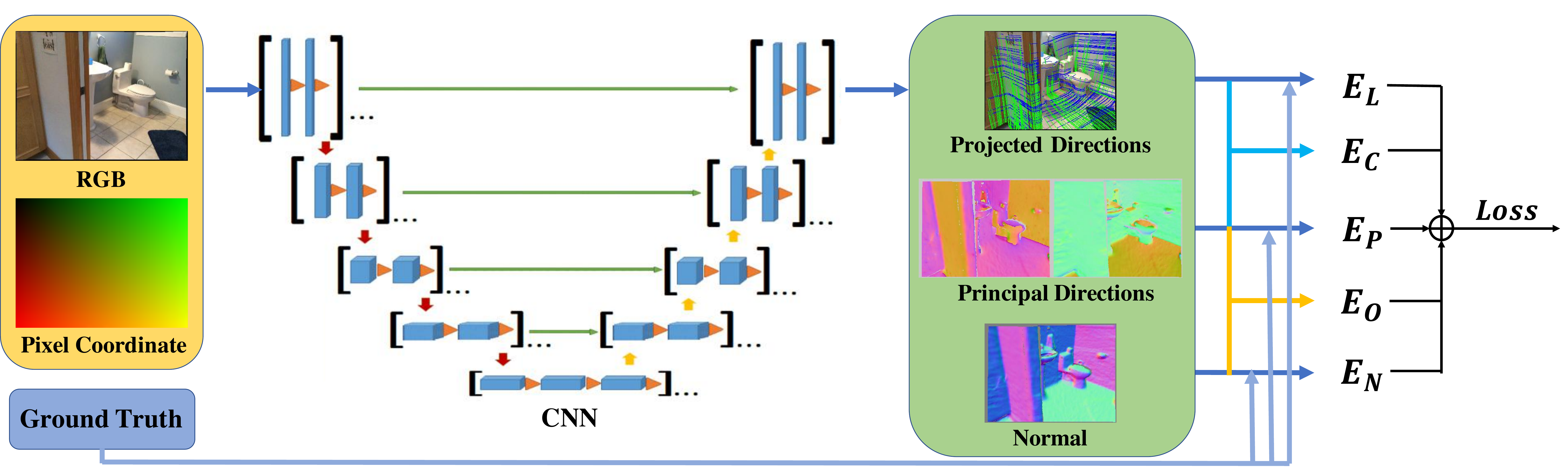}
    \caption{To estimate the local \cframe{}, we feed the RGB image and the canonical pixel coordinate map to the network. The output is a 13-dimensional vector for each pixel including two projected tangent principal directions, two 3D tangent principal directions, and one normal vector. We propose a new loss that utilizes the projected directions to improve the estimation of the \cframe{}.}
    \label{fig:architecture}
\vspace{-0.1in}
\end{figure*}

\subsection{Projected Principal Directions}
\label{sec:project}
Since we aim to predict 3D principal tangent directions from their appearances into RGB images, we first derive the projective geometry that relates them.


 For a pixel $\mb{p}=(p_x,p_y)$ in the canonical camera coordinate system, its 3D position of the pixel can be represented as $\mb{P}=(p_xd, p_yd, d)$ where $d$ is the depth value. Suppose the pixel has two tangent principal directions $\mb{i}$ and $\mb{j}$, and we want to analyze their projections. For $\mb{i}=(i_x,i_y,i_z)$, we can project a line segment $l(\mb{P},\delta,\mb{i})$ that connects endpoints $\mb{P}$ and $\mb{P}+\delta \cdot \mb{i}$ into the image as $l_p(\textbf{P},\delta,\mb{i})$, which is the offset from $\mb{p}$ to the projection of $\mb{P}+\delta \mb{i}$:
\begin{equation}
    l_p(\textbf{P},\delta,\textbf{i}) = \frac{\textbf{P}+\delta \textbf{i}}{(\textbf{P}+\delta \textbf{i})_z}-\mb{p}=(i_x - p_x i_z, i_y - p_y i_z)\frac{\delta}{d+\delta i_z}
\end{equation}

\begin{figure}
    \centering
     \includegraphics[width=0.8\linewidth]{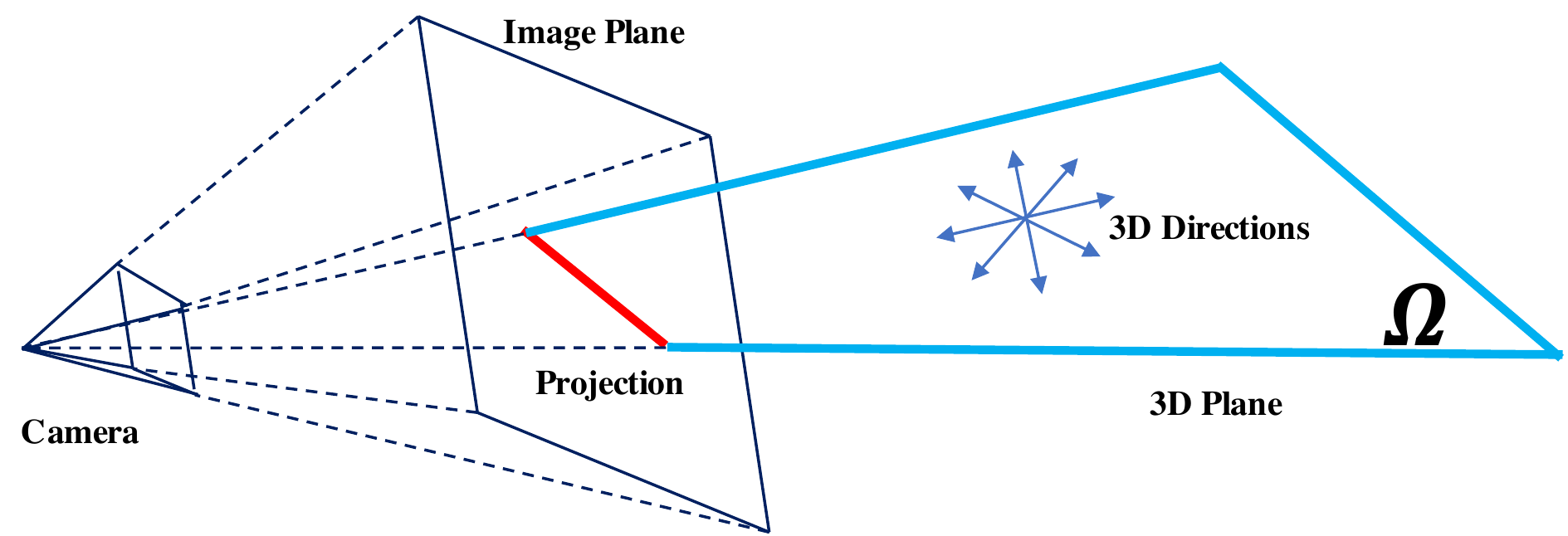}
     \caption{Each projected direction in the image plane (shown in red) corresponds to a 3D plane $\Omega$ in the scene. Any 3D direction inside the plane is a valid candidate for this direction.}
    \label{fig:dir-constraint}
\vspace{-0.2in}
\end{figure}

We find several ways to translate the projected line segment as a property of the pixel, as shown in equation~\ref{eq:def1},\ref{eq:def2},\ref{eq:def3}. The most straightforward idea is to define the property as the projection of the unit 3D line segment from the pixel through the principal directions, represented as 
\begin{equation}
    l^1_p(\textbf{P},\textbf{i}) := l_p(\textbf{P},1,\mb{i}).
    \label{eq:def1}
\end{equation}
This simple definition, however, requires a complex mathematics form including the depth value as a hidden information. Thus it could be hard to learn. Another property is the normalized projected principal direction, or
\begin{equation}
    l^u_p(\textbf{P},\textbf{i}) := \frac{l_p(\textbf{P},\delta,\mb{i})}{||l_p(\textbf{P},\delta,\mb{i})||_2} = \frac{(i_x - p_x i_z, i_y - p_y i_z)}{||(i_x - p_x i_z, i_y - p_y i_z)||_2}.
    \label{eq:def2}
\end{equation}
This representation removes the influence of depth as the challenging hidden property. Since the projection usually aligns with the image gradients, it can be as easy as the task of predicting the normalized gradient for the neural network. However, though this is an easy task, the unit projected direction cannot determine the original 3D direction. As shown in figure~\ref{fig:dir-constraint}(a), a 2D direction in an image is corresponding to a plane in the 3D world, in which any 3D direction could be a valid solution. Fortunately, we can simplify the definition as
\begin{equation}
    l_p^*(\textbf{P},\textbf{i}) := (i_x - p_xi_z, i_y-p_yi_z).
    \label{eq:def3}
\end{equation}
This excludes the influence of the depth and gives enough supervision to the directions in the 3D space. Mathematically, given the prediction of $\mb{l}_p^*(\mb{P},\mb{i})=(l^\mb{i}_x,l^\mb{i}_y)$, we can compute direction $\mb{i}=(i_x,i_y,i_z)$ by solving the system~\ref{eq:solve}.
\begin{equation}
\begin{cases}
  i_x - p_x i_z = l^\mb{i}_x\\
  i_y - p_y i_z = l^\mb{i}_y\\
  i_x^2 +i_y^2 + i_z^2 = 1
\end{cases}
\label{eq:solve}
\end{equation}
    \vspace{-0.1in}
\subsection{Joint Estimation}
\label{sec:network}
We could train a network to estimate the projected principal directions $\mb{i}_p=l_p^*(\textbf{P},\textbf{i})$ and $\mb{j}_p=l_p^*(\textbf{P},\textbf{j})$, and directly infer $\mb{i}$ and $\mb{j}$ according to equation~\ref{eq:solve} for \cframe{} estimation. However, we find that this approach does not lead to a robust \cframe{}. Therefore, we propose to jointly estimate the \cframe{} as well as the projected tangent principal directions, and enforce their orthogonality and projection consistency with additional soft energy constraints.  We expect that the extra constraints will provide a regularization that can help the network learn.

Our proposed solution is illustrated in figure~\ref{fig:architecture}. The neural network can be viewed as a black box function that predicts per-pixel features for the RGB image. Since the projected tangent principal directions relate to the pixel coordinate in the canonical camera, we feed the canonical pixel coordinate together with its RGB values into the network as the input. The network outputs a 13-dimensional vector includes two tangent principal directions $\mb{i}$ and $\mb{j}$, their 2D projections $\mb{i}_p$ and $\mb{j}_p$, and the surface normal $\mb{n}$.

We propose a set of energies so that projected tangent principal directions can assist the local principal axes estimation. The loss energy $E$ is a linear combination of five energy terms as shown in equation~\ref{eq:loss},
\begin{align}
\begin{split}
  E = \lambda_L &E_L + \lambda_P E_P + \lambda_N E_N + \lambda_C E_C + \lambda_O E_O\\
  E_L &= \min_{0\leq k \le 4} ||[\mb{i}_p,\mb{j}_p] - R_k([\mb{i}_p^{gt},\mb{j}_p^{gt}])||_2^2\\
  E_P &= \min_{0\leq k \le 4} ||[\mb{i},\mb{j}] - R_k([\mb{i}^{gt},\mb{j}^{gt}])||_2^2\\
  E_N &= ||N - N^{gt}||_2^2\\
  E_C &= ||l^*_p(\mb{i}) - \mb{i}_p||_2^2+||l^*_p(\mb{j}) - \mb{j}_p||_2^2\\
  E_O &= ||N - \mb{i}\times\mb{j}||_2^2 \\
\end{split}
\label{eq:loss}
\end{align}
where $R_1([\mb{a},\mb{b}])=[-\mb{b},\mb{a}]$ and $R_k = R_1\circ R_{k-1} (k>1)$.

Specifically, $E_L$ measures the distance between the predicted tangent principal directions and the ground truth in the 2D projected space. $R_k$ represents the $90k$ degree rotation around the normal axis. $E_L$ removes the rotational ambiguity by enumerating the possible $90k^{\circ}$ rotations and measure the minimum L2 loss among them. Similarly, $E_P$ measures the minimum L2 loss of tangent principal directions in the 3D space, and $E_N$ measures the L2 loss of the surface normal estimation. In order to connect the tangent principal directions to their projections, we design $E_C$ to measure the consistency between the projected predicted directions ($l^*_p(\mb{i})$,$l^*_p(\mb{j})$) and the predicted one ($\mb{i}_p$,$\mb{j}_p$) by the network. Finally, we also hope the influence can be propagated to the surface normal, so we add an orthogonality constraint $E_O$ to enforce that the surface normal is orthogonal to the tangent principal directions.

Since all the distances are roughly on the same scale, we set $\lambda_L=\lambda_P=\lambda_N=1$ to balance the penalty for errors for different vectors. To enforce the system to predict orthogonal \cframe{} with consistent 2D projection, we set $\lambda_C=\lambda_O=5$ in our experiments to provide slightly stronger constraints between network predictions.
\section{Evaluation}
\label{sec:evaluation}
In this section, we describe a series of experiments to evaluate our method for local \cframe{} estimation and do ablation studies using the ScanNet dataset~\cite{dai2017scannet}. 
Unless otherwise specified, we used the DORN architecture~\cite{fu2018deep} as the backbone for the architecture in fig. \ref{fig:architecture}, and we used equation~\ref{eq:def3} for the projected tangent principal directions, since they gave the best results (see below).
The main conclusion of these tests is that jointly predicting the projected tangent directions and enforcing the consistency loss are major contributors to the success of local principal axes and surface normal estimation.

\begin{figure}[t]
    \centering
    \includegraphics[width=\linewidth]{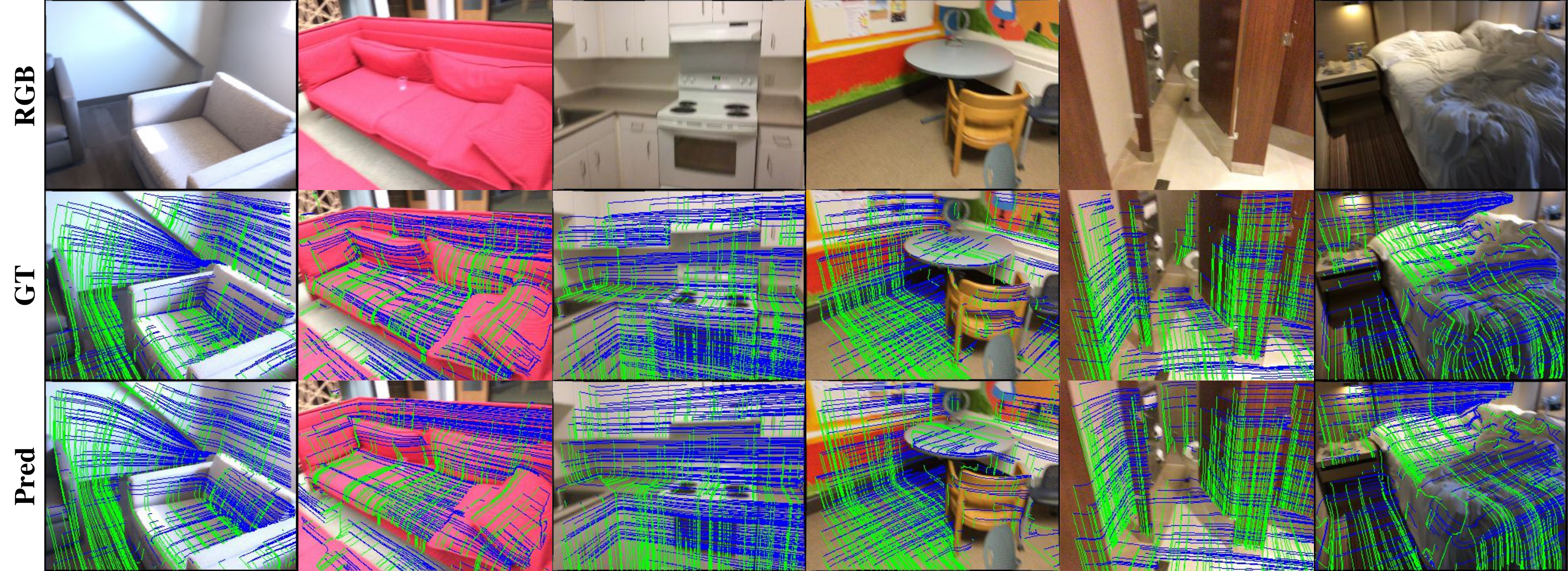}
    \caption{Visualization of the projected principal directions. Our estimation is similar to the ground truth at curved surfaces or texture smooth regions. The predicted directions align with textures and gradients in the input image.}
    \label{fig:project}
\end{figure}

\vspace{-0.1in}
\paragraph{How well can canonical frames be estimated from RGB?}  Our first experiment simply investigates how well our algorithm can predict the \cframe{}.   Since this is a new task, there is no suitable comparison to prior work.   However, we can still gain insight into the problem by comparing errors in predicted normals, principal tangent principal directions, and projected tangent principal directions.   The results in table~\ref{tab:3dframe} show that prediction of projected tangent principal directions have least error, surface normals have most error, and tangent principal directions are in the middle.   This suggests that predicting tangent directions is less error prone than normals, which should be expected since they largely align with textures and gradients in the input image (figure~\ref{fig:project}). 

\begin{table}[t]
    \centering
    \small
    \tabcolsep=0.12cm
    \begin{tabular}{|c|c|c|c||c|c|c|}
        \hline
         \textbf{3D Frame} & mean & median & rmse & $11.25^\circ$ & $22.5^\circ$ & $30^\circ$\\
         \hline
         Normal & 15.28 & 8.14 & 23.36 & 60.6 & 78.6 & 84.7\\
         \hline
         Principal & 12.26 & 7.88 & 16.85 & 63.7 & 84.3 & 90.8\\
         \hline
         Projection & 7.55 & 4.46 & 11.36 & 79.8 & 93.0 & 96.3\\
         \hline
    \end{tabular}
    \caption{Testing mean average error of local principal axes estimation on ScanNet~\cite{dai2017scannet}. We evaluate surface normals, tangent principal directions their projections predicted by our network.}
    \label{tab:3dframe}
\end{table}

\begin{table}
    \centering
    \small
    \begin{tabular}{|c|c|c|c|c|c|}
        \hline
         Method & UNet & SkipNet & GeoNet & DORN\\
         \hline
         Normal & 21.08 & 20.84 & 20.37 & 16.42\\
         \hline
         Normal-YZ & 17.49 & 17.17 & 16.71 & 12.51\\
         \hline
         Normal-XZ & 18.05 & 17.16 & 17.68 & 13.00\\
         \hline
         Normal-XY & 29.05 & 29.71 & 29.08 & 22.57\\
         \hline
         Principal & 17.55 & 15.78 & 15.41 & 12.53\\
         \hline
         Principal-YZ & 21.15 & 21.96 & 20.61 & 16.19\\
         \hline
         Principal-XZ & 22.67 & 21.87 & 21.57 & 16.65\\
         \hline
         Principal-XY & 11.47 & 9.96 & 9.53 & 7.55\\
         \hline
    \end{tabular}
    \caption{Mean angle errors of normals and tangent principal directions and their projections to three orthogonal planes on ScanNet.}
    \label{tab:error}
\end{table}
\label{sec:ablate}

\begin{figure}
    \centering
    \includegraphics[width=0.75\linewidth]{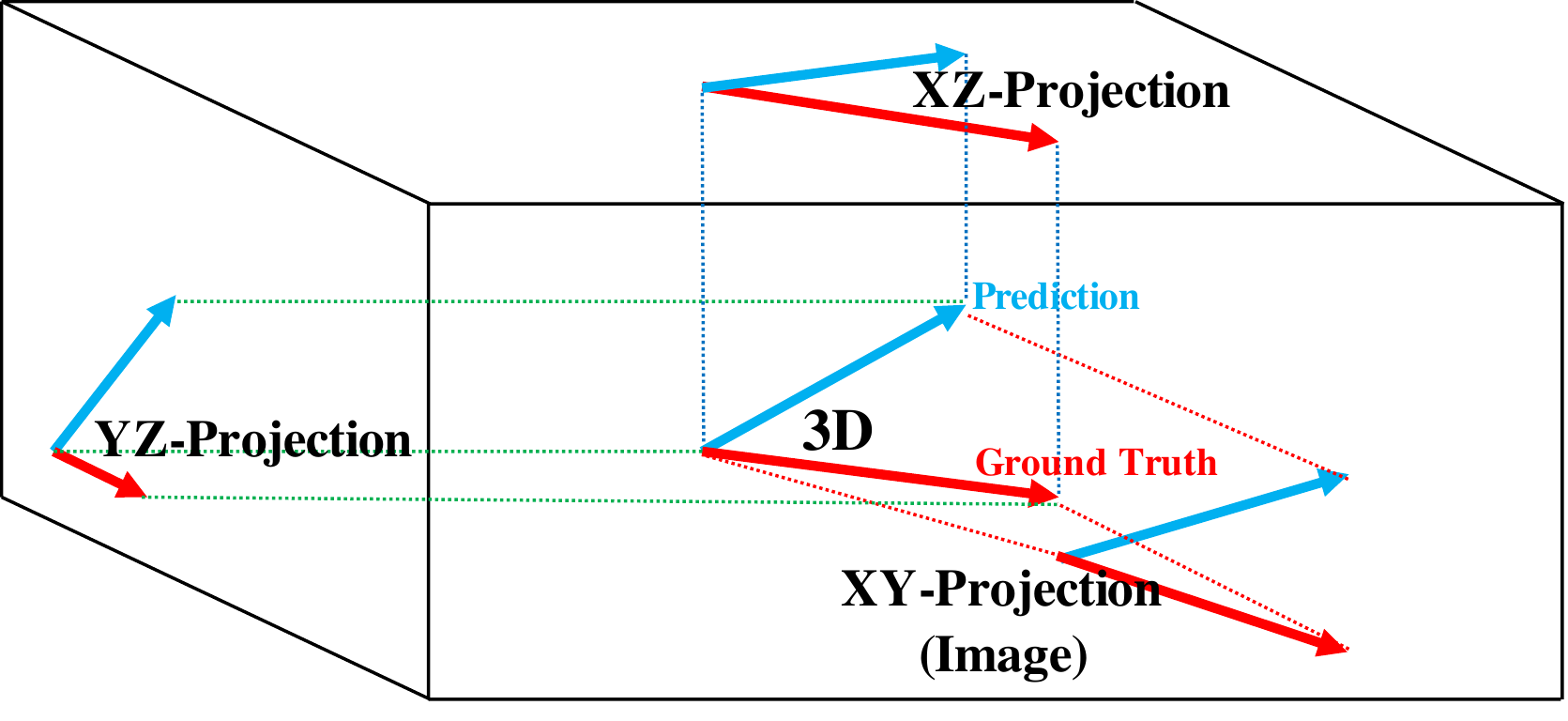}
    \caption{By projecting the directions into XY, YZ, XZ planes in the camera space, we can measure the projected angle error.}
    \label{fig:err-proj}
\end{figure}

\vspace{-0.1in}
\paragraph{Which frame directions are easiest to predict?}  To further investigate the relative challenge of predicting different components of the local \cframe{}, we perform experiments in which we separately train normals and tangent principal directions in 3D space with L2 losses and evaluate them with mean angle errors of their projections to three planes in camera space, as illustrated in figure~\ref{fig:err-proj}.  The prediction errors and their projected components, listed in table~\ref{tab:error}, suggest
that the errors of the tangent principal directions are less than those of normals, and the projected errors on the image plane are smaller than those on the other two planes for tangent principal directions.  This again suggests that the network can predict tangent principal directions better than surface normals, especially for the components projected into the image plane.  Interestingly, the projected errors for the normal in the image plane is the largest, which might be because the network learns tangent principal directions in the latent space and propagates the errors from XZ and YZ planes to the image plane by the cross product.

\vspace{-0.1in}
\paragraph{How does each loss contributes to the estimation?}
We next study how our proposed consistency losses influence the learning process. In table~\ref{tab:consistency}, we present the testing mean average angle for surface normals w/o. certain parts of losses during training on ScanNet. We note that by directly predicting all $E_N$ and $E_P$ together, there is already an improvement. The reason could be that the correlation between predicted principal directions and the 3D frames are automatically learned from the data distribution. However, the improvement is minor without predicting the projected principal directions with $E_L$. With orthogonal or consistency constraints, the performance can be further improved and achieve maximum with both.
\begin{table}
    \centering
    \tabcolsep=0.20cm
    \small
    \begin{tabular}{|c|c|c|c|c|c|}
        \hline
         Method & UNet & SkipNet & GeoNet & DORN\\
         \hline
         $E_N$ & 21.08 & 20.36 & 19.77 & 16.42\\
         \hline
         $E_N$,$E_P$ & 21.04 & 20.45 & 19.64 & 16.29\\
         \hline
         $E_N$,$E_P$,$E_L$ & 20.62 & 19.47 & 19.26 & 15.45\\
         \hline
         $E_N$,$E_P$,$E_L$,$E_O$ & 20.58 & 19.43 & 19.18 & 15.41\\
         \hline
         $E_N$,$E_P$,$E_L$,$E_C$ & 19.79 & 19.44 & 19.02 & 15.31\\
         \hline
         All Losses & \textbf{19.68} & \textbf{19.39} & \textbf{18.96} & \textbf{15.28}\\
         \hline
    \end{tabular}
    \caption{We test mean average angle errors for surface normal predictions with different combination of loss terms on ScanNet. $E_L$ and $E_C$ has major contributions to the improvement, suggesting the importance of the projected principal directions.}
    \label{tab:consistency}
    \vspace{-0.1in}
\end{table}

\vspace{-0.1in}
\paragraph{Does the method generalize to different networks?}  To study the generality of our approach, we tested it with different network architectures.   Table~\ref{tab:consistency} shows that our joint losses improve performance for all the tested networks including UNet\cite{ronneberger2015u}, SkipNet\cite{bansal2016marr}, GeoNet\cite{qi2018geonet} and DORN\cite{fu2018deep}.

\vspace{-0.1in}
\paragraph{Which definition of projected directions is best?}
In equation~\ref{eq:def1}~\ref{eq:def2}~\ref{eq:def3}, we propose three choices for projected tangent principal directions. We use UNet~\cite{ronneberger2015u} to separately train and test them on ScanNet~\cite{dai2017scannet} as shown in table~\ref{tab:def}. The mean angle error for equation~\ref{eq:def1} is the highest as a complex function related to the depth. The error for equation~\ref{eq:def3} is only slightly higher than that in equation~\ref{eq:def2}, but equation~\ref{eq:def3} can explicitly guide the 3D directions with the consistency loss $E_C$. Therefore, we select equation~\ref{eq:def3} together with the canonical frames for joint estimation.

\begin{table}[t]
    \centering
    \tabcolsep=0.13cm
    \small
    \begin{tabular}{|c|c|c|c||c|c|c|}
        \hline
         \textbf{ScanNet} & mean & median & rmse & $11.25^\circ$ & $22.5^\circ$ & $30^\circ$\\
         \hline
         $l^1_p(\mb{P},\mb{i})$ & 11.13 & 7.63 & 15.00 & 65.1 & 86.2 & 92.5\\
         \hline
         $l^u_p(\mb{P},\mb{i})$ & \textbf{7.35} & \textbf{4.38} & \textbf{10.94} & \textbf{81.2} & \textbf{93.6} & \textbf{96.7}\\
         \hline
         $l^*_p(\mb{P},\mb{i})$ & 7.56 & 4.46 & 11.36 & 79.8 & 93.0 & 96.3\\
         \hline
    \end{tabular}
    \caption{Testing mean average error of different choices for projected tangent principal directions on ScanNet dataset.}
    \label{tab:def}
\end{table}

\section{Applications}
\label{sec:applications}

In this section, we investigate whether the estimation of local \cframe{} is useful for applications.   We first study surface normal estimation, a direct application of our method.   In addition, we study how 3D \cframe{} can be utilized for perspective invariant feature descriptors and augmented reality.

\subsection{Surface Normal Estimation}
\label{sec:normal}

\paragraph{Test on ScanNet}
We first compare the performance of our surface normal estimation with state-of-the-art methods on ScanNet~\cite{dai2017scannet}. 
We use our approach to train four networks and evaluate them according to ground truth provided by RGBD. Table~\ref{tab:scannet-comparison} shows the results for all networks including UNet~\cite{ronneberger2015u}, SkipNet~\cite{bansal2016marr}, GeoNet~\cite{qi2018geonet} and DORN~\cite{fu2018deep}. With the assistance of the projected tangent principal directions, the normal prediction is better for all architectures.

\begin{table}
    \centering
    \tabcolsep=0.08cm
    \small
    \begin{tabular}{|c|c|c|c||c|c|c|}
        \hline
         \textbf{ScanNet} & mean & median & rmse & $11.25^\circ$ & $22.5^\circ$ & $30^\circ$\\
         \hline
         UNet & 21.08 & 14.21 & 28.55 & 40.8 & 66.9 & 76.3\\
         \hline
         UNet-Ours & 19.68 & 12.43 & 27.58 & 46.1 & 70.6 & 78.8\\
         \hline
         SkipNet & 20.36 & 13.74 & 28.63 & 45.4 & 68.2 & 77.4\\
         \hline
         SkipNet-Ours & 19.39 & 10.85 & 27.52 & 53.2 & 72.7 & 79.3\\
         \hline
         GeoNet & 19.77 & 11.34 & 28.51 & 49.7 & 70.4 & 77.7\\
         \hline
         GeoNet-Ours & 18.96 & 9.84 & 27.29 & 54.6 & 73.5 & 80.1\\
         \hline
         DORN & 16.42 & 8.64 & 24.94 & 58.7 & 76.7 & 82.9\\
         \hline
         DORN-Ours & \textbf{15.28} & \textbf{8.14} & \textbf{23.36} & \textbf{60.6} & \textbf{78.6} & \textbf{84.7}\\
         \hline         
    \end{tabular}
    \caption{Evaluation on Surface Normal Predictions. We train and test our algorithm with different network architectures on the ScanNet~\cite{dai2017scannet} dataset. Assisted by our joint loss, the performances of all networks are improved.}
    \label{tab:scannet-comparison}
    \vspace{-0.1in}
\end{table}

Figure~\ref{fig:vis-result} visualizes the normals predicted using DORN with and without our method. With our approach, the errors are smaller especially at object boundaries, possibly because of the additional supervision given by the projected tangent principal directions.
\begin{figure}
    \centering
    \includegraphics[width=\linewidth]{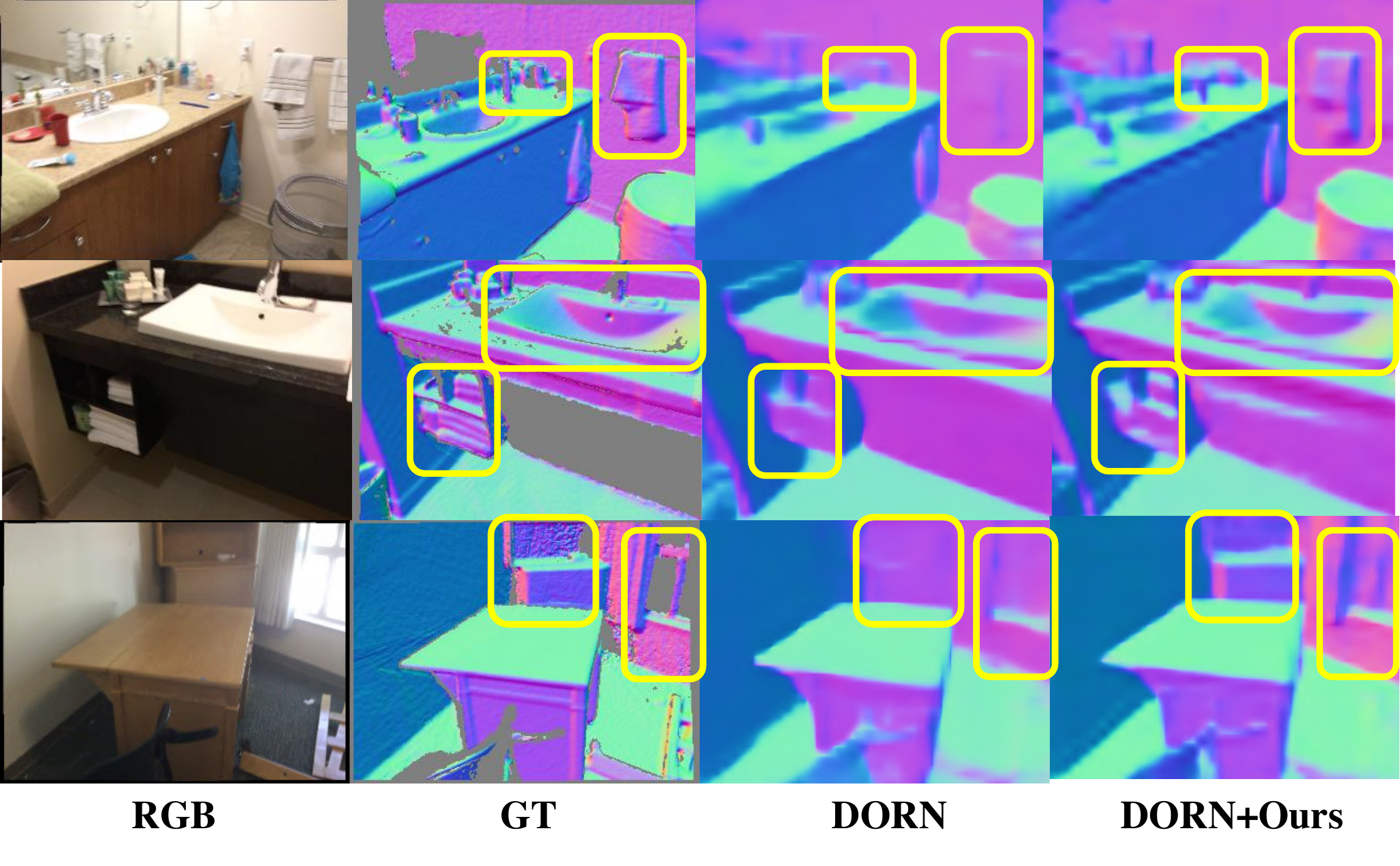}
    \vspace{-0.2in}
    \caption{Visual comparison of the results. With our joint loss, the predicted surface normals produce less errors and more details.}
    \label{fig:vis-result}
\end{figure}

\vspace{-0.1in}
\paragraph{Test on NYUv2}
\label{sec:transfer}
We test different versions of our network on NYUv2~\cite{eigen2014depth} as a standard evaluation dataset. Since NYUv2 does not provide reconstructed 3D meshes, we cannot get ground truth 3D frames. Therefore, we train the network on ScanNet datasets and directly test on NYUv2, as shown in Table~\ref{tab:vis-nyu}. Note that GeoNet-origin~\cite{qi2018geonet} is specifically trained and tested on NYUv2 and is the current state-of-the-art method on normal estimation for that dataset. Other rows are networks trained with and without our joint losses on ScanNet and tested on NYUv2. 

\begin{table}
    \centering
    \tabcolsep=0.07cm
    \small
    \begin{tabular}{|c|c|c|c||c|c|c|}
        \hline
         \textbf{NYUv2} & mean & median & rmse & $11.25^\circ$ & $22.5^\circ$ & $30^\circ$\\
         \hline
         GeoNet-origin & 19.0 & 11.8 & 26.9 & 48.4 & 71.5 & \textbf{79.5}\\
         \hline
         \hline
         \textbf{ScanNet} & mean & median & rmse & $11.25^\circ$ & $22.5^\circ$ & $30^\circ$\\
         \hline
         UNet & 23.46 & 17.58 & 29.90 & 29.9 & 60.9 & 72.7\\
         \hline
         UNet-Ours & 22.09 & 15.45 & 29.26 & 36.9 & 64.5 & 74.9\\
         \hline
         SkipNet & 22.27 & 14.25 & 30.60 & 42.0 & 64.8 & 73.5\\
         \hline
         SkipNet-Ours & 20.68 & 13.42 & 28.33 & 46.3 & 67.4 & 76.0\\
         \hline
         GeoNet & 22.02 & 14.55 & 29.79 & 40.7 & 64.9 & 73.9\\
         \hline
         GeoNet-Ours & 20.22 & 13.23 & 28.19 & 47.9 & 68.0 & 76.4\\
         \hline
         DORN & 19.12 & 11.60 & 27.06 & 49.0 & 70.6 & 78.5\\
         \hline
         DORN-Ours & \textbf{18.63} & \textbf{11.16} & \textbf{26.61} & \textbf{50.2} & \textbf{71.6} & \textbf{79.5}\\
         \hline         
    \end{tabular}
    \caption{Normal prediction on NYUv2~\cite{eigen2014depth}. GeoNet-origin trained and tested on NYUv2~\cite{qi2018geonet}.  DORN-Ours trained on ScanNet performs best among all.}
    \label{tab:vis-nyu}
\vspace{-0.1in}
\end{table}

Although GeoNet performs worse than GeoNet-origin by training only on ScanNet without fine-tuning, we still achieve better performance with the DORN~\cite{fu2018deep} architecture and our loss (DORN-Ours). Moreover, all networks show better performance with our loss, implying a robust advantage of our joint estimation. 

\subsection{Keypoint Matching}

Predicting local transformations is important for 
keypoint feature matching~\cite{lowe2004distinctive,bay2006surf,tola2010daisy,han2015matchnet,zagoruyko2015learning,simo2015discriminative,yi2016lift}. 
For example, SIFT~\cite{lowe2004distinctive} estimates scale and camera-plane rotations to provide invariance to those transformations.  Since
our network estimates a full local 3D \cframe{}, we can additionally estimate a projective warp.  
Specifically, predicting the pairs of projected tangent principal directions (in equation~\ref{eq:def3}) for pixel $\mb{p}$ as $\mb{i}_p$ and $\mb{j}_p$ the local patch $\mathbb{P}$ is warped to $\mathbb{P}^*$ as shown in equation~\ref{eq:warp}.
\begin{equation}
    \mathbb{P}^*(\textbf{x}) = \mathbb{P}([\textbf{i}_p, \textbf{j}_p]\textbf{x})
    \label{eq:warp}
\end{equation}

\begin{figure}[t]
\centering
    \includegraphics[width=0.8\linewidth]{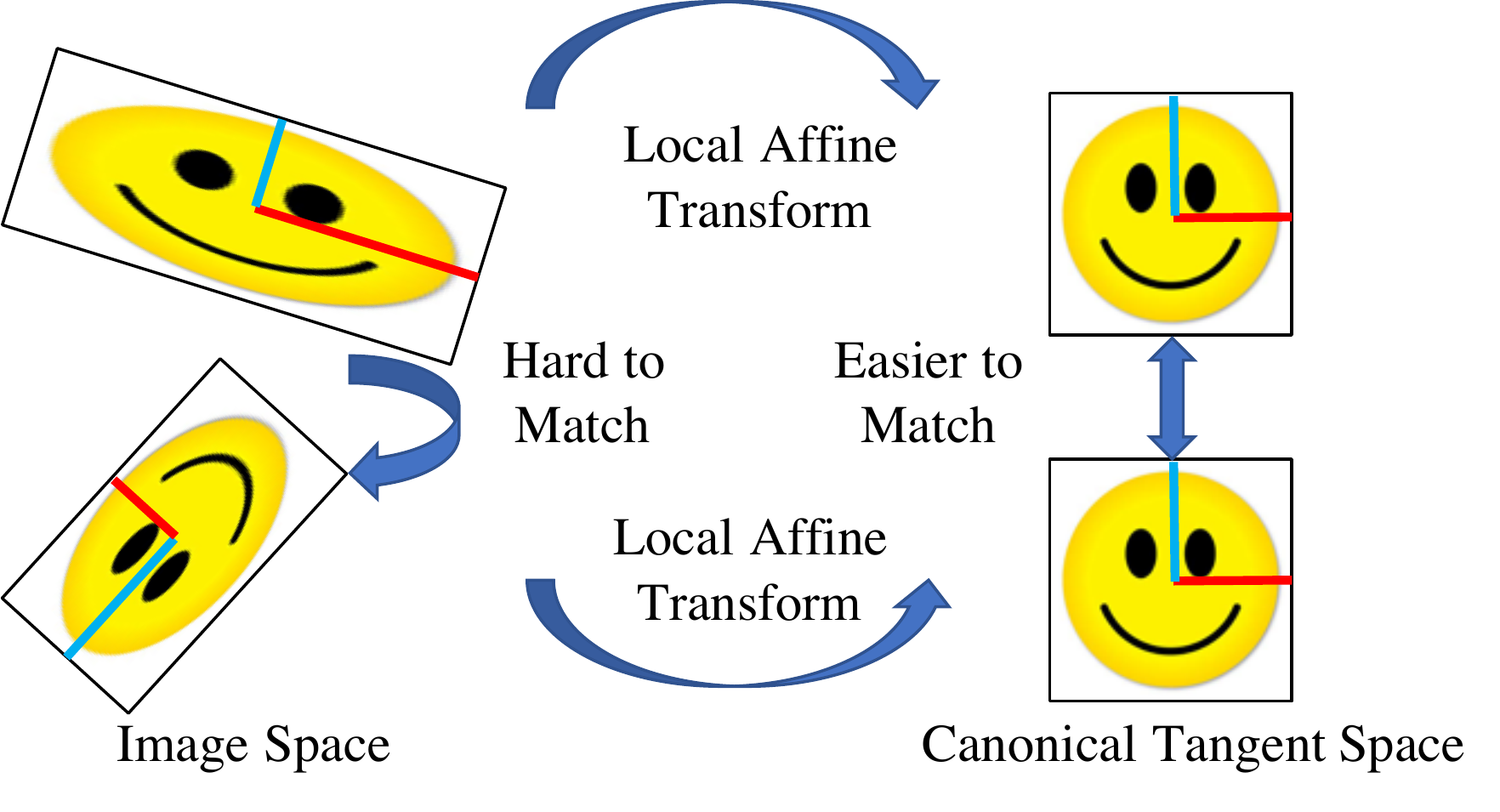}
    \caption{By warping the local patch from the image to the canonical tangent plane of the surface, feature descriptors are invariant to the camera perspectives. Keypoint matching could be improved.}
    \label{fig:warp}
\end{figure}

To investigate this feature, we performed a simple experiment with SIFT~\cite{lowe2004distinctive}.  We augmented the standard SIFT descriptor computation to account for perspective warps implied by our predicted \cframes. Specifically, we detect keypoints using SIFT~\cite{lowe2004distinctive}, and extract the SIFT descriptors on the warped patch using our estimated local projected tangent principal directions.

\begin{figure}
    \centering
    \includegraphics[width=\linewidth]{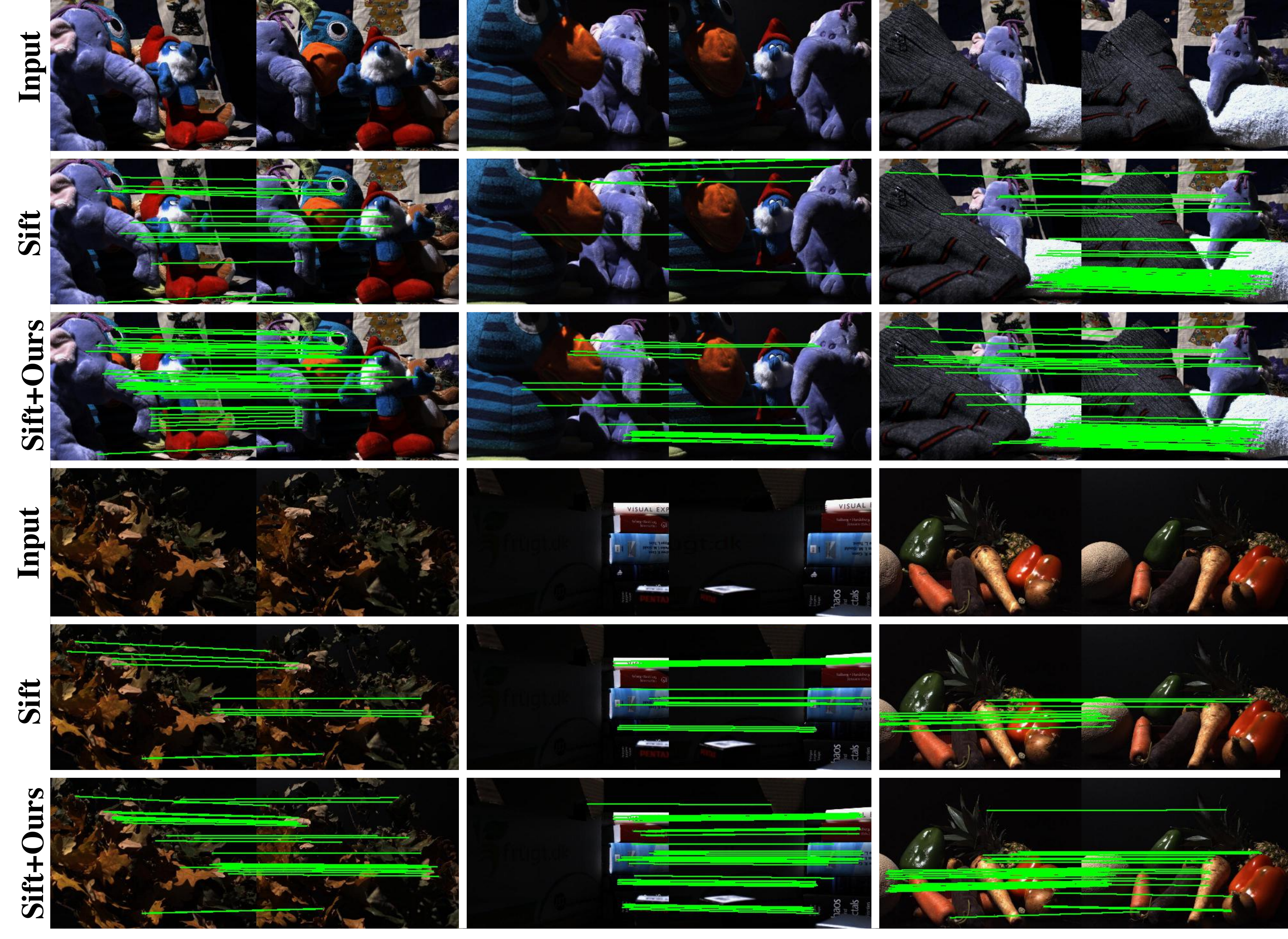}
    \caption{Visualize the matching between SIFT with and without our warping. With our warping, SIFT finds more correct matches.}
    \label{fig:dtu-vis}
    \vspace{-0.1in}
\end{figure}

To evaluate our modified descriptor, we compare it with other methods on the DTU dataset~\cite{aanaes2012interesting}, where scenes are captured with different lighting and viewpoints. We visualize the correct matching produced by SIFT with and without our local image warping in figure~\ref{fig:dtu-vis}. As a result, the local image warping reduces the perspective distortions and produce more correct matches.
We also test the matching score as ``the ratio of ground truth correspondences that can be recovered by the whole pipeline over the number of features proposed by the pipeline in the shared viewpoint region''~\cite{yi2016lift}.
\begin{table}
    \centering
    \tabcolsep=0.12cm
    \begin{tabular}{|c|c|c|c|}
        \hline
         SURF~\cite{bay2006surf} & ORB~\cite{rublee2011orb} & Daisy~\cite{tola2010daisy} & BRISK~\cite{leutenegger2011brisk}\\
         \hline
         .224 & .127 & .262 & .193\\
        \hline
         VGG~\cite{kahler2015very} & MatchNet~\cite{han2015matchnet} & DeepDesc~\cite{simo2015discriminative} & PN-Net~\cite{balntas2016pn}\\
         \hline
         .271 & .198 & .257 & .267\\
        \hline
         SIFT~\cite{lowe2004distinctive} & ASIFT~\cite{yu2011asift} & LIFT~\cite{yi2016lift} & SIFT+Ours\\
         \hline
         .272 & .265 & .317 & .335\\
        \hline
    \end{tabular}
    \caption{Matching score of descriptors on the DTU dataset.}
    \label{tab:matching}
\vspace{-0.2in}
\end{table}
As shown in table~\ref{tab:matching}, SIFT~\cite{lowe2004distinctive} outperforms most methods. Since our method additionally reduces the perspective effects using the projected tangent principal directions, we can further improve the SIFT performance. Note that ASIFT~\cite{yu2011asift} also perceives the limitation of SIFT~\cite{lowe2004distinctive} to different viewpoints, and extracts keypoints from the image with various affine transforms. Therefore, they usually provide many more correct matching but also more outliers. That is why the matching score produced by ASIFT~\cite{yu2011asift} is slightly lower than SIFT~\cite{lowe2004distinctive}. However, it sometimes shows better robustness assisted by geometric filters in certain applications.

\subsection{Augmented Reality}
A particularly compelling application of predicting 3D surface frames is augmented reality -- i.e., it enables adding new elements to a scene with appropriate 3D orientations.

\begin{figure}
    \centering
    \includegraphics[width=\linewidth]{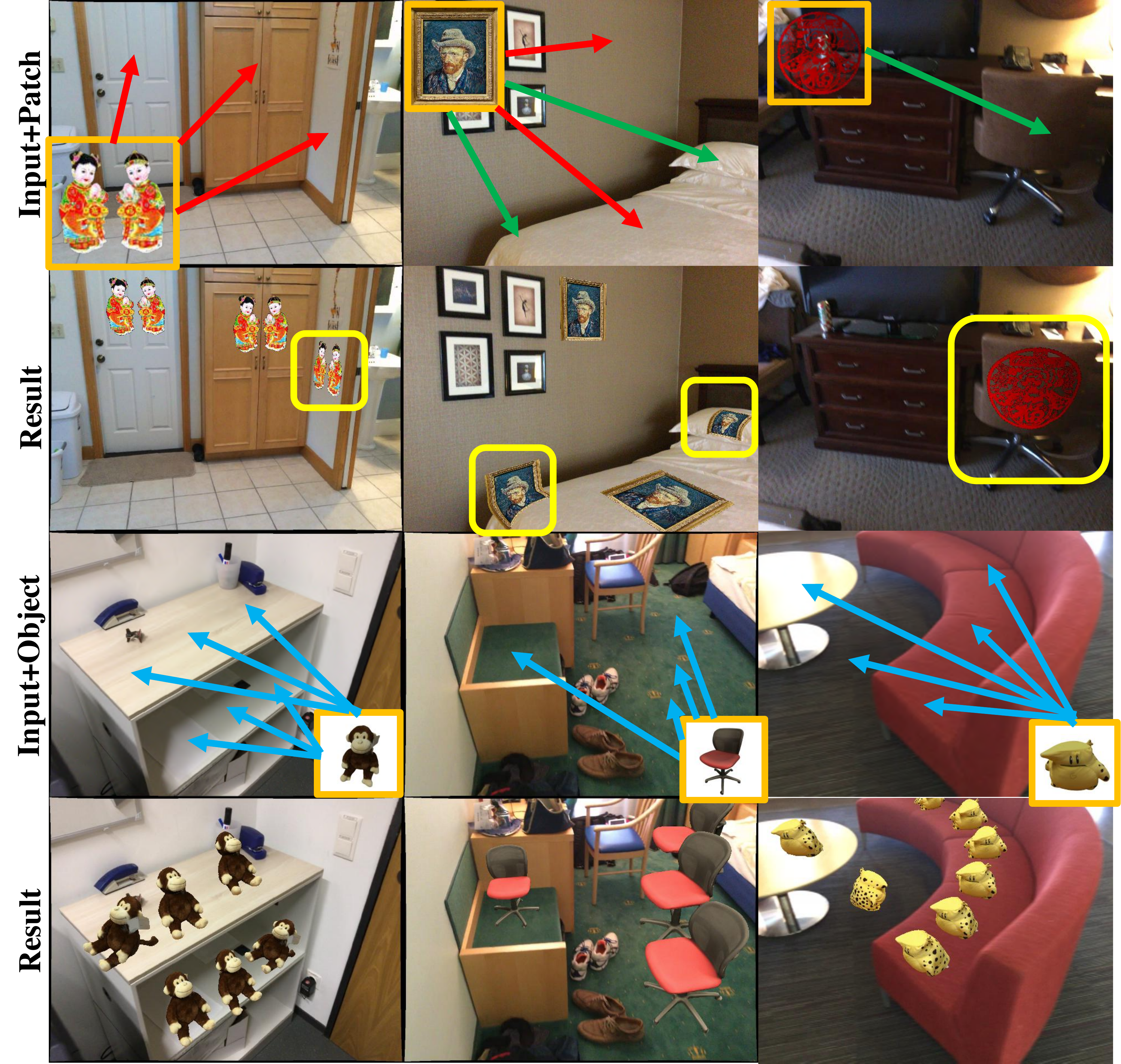}
    \caption{Adding new elements in the scene. We  use red arrows to represent rigid attachment, green to represent deformable attachment, and blue to represent object placement.}
    \label{fig:attach}
    \vspace{-0.15in}
\end{figure}

\vspace{-0.1in}
\paragraph{Decal Attachment}  As a simple example, we investigate warping virtual decals added to RGB images based on the estimated 3D frame (first two rows of figure~\ref{fig:attach}). In our experiment, we ask the user to select one pixel in an RGB image to indicate the center point for the decal on a surface.  If we assume the surface is planar, we can compute the homography transformation required to align the decal with the scene geometry. Suppose the selected pixel is $\mb{p}$ with two estimated principal directions $\mb{i}$ and $\mb{j}$ and depth $d$. Then, the center of the pattern $(x_c,y_c)$ is located at $K^{-1}\mb{p}\cdot d$ where $K$ is the camera intrinsics. We additionally suppose that the target distance of neighboring pixels of the pattern attached to the scene is $\delta\cdot d$. Then, for pixel $(x,y)$ in the pattern, the homogeneous coordinate in the scene is
\begin{equation}
\textbf{P}(x,y) = K\cdot(K^{-1}\mb{p}\cdot d + \textbf{i}\cdot (x-x_c)\delta d + \textbf{j}\cdot (y-y_c)\delta d)
\end{equation}
Therefore, the homography transform can be inferred as
\begin{equation}
    H=K[\delta \mb{i}, \delta \mb{j}, K^{-1}\mb{p} - \delta(x_c\mb{i}+y_c\mb{j})]
\end{equation}
Here, $\delta$ represents the relative scale of the pattern to the depth of the pixel, which can be controlled by the user.
Beyond this point, our local frame even enables deformable pattern attachment on curved surfaces. Similarly, the homogeneous coordinate of any pixel $\mb{x}_t$ can be computed as
\begin{equation}
    \mb{P}(\mb{x}_t)=\mb{p} + \delta K \cdot\int_{\mb{x}_c}^{\mb{x}_t} [\textbf{i}(\mb{P}(\mb{x})),\textbf{j}(\mb{P}(\mb{x}))] d\mb{x}
\end{equation}
We use the simple explicit Euler method to evolve $\mb{P}(\mb{x})$, where the path of the integration starts from the center, and follows the order guided by the breadth first search, where the expansion is from one pixel to those among its four neighbors which are not yet visited. Several examples of deformable attachment is shown in figure~\ref{fig:attach}. The user can control $\delta$ to specify the size of the attached patterns.

\vspace{-0.1in}
\paragraph{Object Placement} We can also use the local 3D frame defined by predicted principal axes to render 3D objects into RGB images, as shown in the last two rows of figure~\ref{fig:attach}.  For this application, predicting the full 3D orientation of the scene geometry is critical, so that objects can be planes not only in accordance with the surface normal, but also in the appropriate rotation around the normal (e.g., so that the front is facing the right way).   For example, the stuffed animals in the bottom left of figure~\ref{fig:attach} would appear unnatural if they were facing the wall. This could also eases mixed reality data augmentation for vision tasks, where existing methods require the depth image for plane detection~\cite{wang2019normalized}.

\section{Conclusion}
We propose the novel problem of densely estimating local 3D \cframe{} from a single RGB image. We formulate the problem as a joint estimation of surface normals, canonical tangent directions, and projected tangent directions.  We find this approach leads to superior performance as compared to previous work on normal estimation and other tasks.  Further study is warranted to investigate what other geometric properties can be predicted from RGB using similar methods and how they can be used in other applications.
{\small
\bibliographystyle{ieee}
\bibliography{egbib}

\begin{thebibliography}{10}\itemsep=-1pt

\bibitem{aanaes2012interesting}
H.~Aan{\ae}s, A.~L. Dahl, and K.~S. Pedersen.
\newblock Interesting interest points.
\newblock {\em International Journal of Computer Vision}, 97(1):18--35, 2012.

\bibitem{balntas2016pn}
V.~Balntas, E.~Johns, L.~Tang, and K.~Mikolajczyk.
\newblock Pn-net: Conjoined triple deep network for learning local image
  descriptors.
\newblock {\em arXiv preprint arXiv:1601.05030}, 2016.

\bibitem{bansal2016marr}
A.~Bansal, B.~Russell, and A.~Gupta.
\newblock Marr revisited: 2d-3d alignment via surface normal prediction.
\newblock In {\em Proceedings of the IEEE conference on computer vision and
  pattern recognition}, pages 5965--5974, 2016.

\bibitem{bay2006surf}
H.~Bay, T.~Tuytelaars, and L.~Van~Gool.
\newblock Surf: Speeded up robust features.
\newblock In {\em European conference on computer vision}, pages 404--417.
  Springer, 2006.

\bibitem{bommes2009mixed}
D.~Bommes, H.~Zimmer, and L.~Kobbelt.
\newblock Mixed-integer quadrangulation.
\newblock In {\em ACM Transactions On Graphics (TOG)}, volume~28, page~77. ACM,
  2009.

\bibitem{boscaini2016learning}
D.~Boscaini, J.~Masci, E.~Rodol{\`a}, and M.~Bronstein.
\newblock Learning shape correspondence with anisotropic convolutional neural
  networks.
\newblock In {\em Advances in Neural Information Processing Systems}, pages
  3189--3197, 2016.

\bibitem{cazals2005estimating}
F.~Cazals and M.~Pouget.
\newblock Estimating differential quantities using polynomial fitting of
  osculating jets.
\newblock {\em Computer Aided Geometric Design}, 22(2):121--146, 2005.

\bibitem{cohen2003restricted}
D.~Cohen-Steiner and J.-M. Morvan.
\newblock Restricted {D}elaunay triangulations and normal cycle.
\newblock In {\em Proceedings of the Nineteenth Annual Symposium on
  Computational Geometry}, pages 312--321. ACM, 2003.

\bibitem{dai2017scannet}
A.~Dai, A.~X. Chang, M.~Savva, M.~Halber, T.~A. Funkhouser, and M.~Nie{\ss}ner.
\newblock Scannet: Richly-annotated 3d reconstructions of indoor scenes.
\newblock In {\em CVPR}, volume~2, page~10, 2017.

\bibitem{eigen2015predicting}
D.~Eigen and R.~Fergus.
\newblock Predicting depth, surface normals and semantic labels with a common
  multi-scale convolutional architecture.
\newblock In {\em Proceedings of the IEEE international conference on computer
  vision}, pages 2650--2658, 2015.

\bibitem{eigen2014depth}
D.~Eigen, C.~Puhrsch, and R.~Fergus.
\newblock Depth map prediction from a single image using a multi-scale deep
  network.
\newblock In {\em Advances in neural information processing systems}, pages
  2366--2374, 2014.

\bibitem{fu2018deep}
H.~Fu, M.~Gong, C.~Wang, K.~Batmanghelich, and D.~Tao.
\newblock Deep ordinal regression network for monocular depth estimation.
\newblock In {\em Proceedings of the IEEE Conference on Computer Vision and
  Pattern Recognition}, pages 2002--2011, 2018.

\bibitem{garg2016unsupervised}
R.~Garg, V.~K. BG, G.~Carneiro, and I.~Reid.
\newblock Unsupervised cnn for single view depth estimation: Geometry to the
  rescue.
\newblock In {\em European Conference on Computer Vision}, pages 740--756.
  Springer, 2016.

\bibitem{geiger2013vision}
A.~Geiger, P.~Lenz, C.~Stiller, and R.~Urtasun.
\newblock Vision meets robotics: The kitti dataset.
\newblock {\em The International Journal of Robotics Research},
  32(11):1231--1237, 2013.

\bibitem{han2015matchnet}
X.~Han, T.~Leung, Y.~Jia, R.~Sukthankar, and A.~C. Berg.
\newblock Matchnet: Unifying feature and metric learning for patch-based
  matching.
\newblock In {\em Proceedings of the IEEE Conference on Computer Vision and
  Pattern Recognition}, pages 3279--3286, 2015.

\bibitem{he2016deep}
K.~He, X.~Zhang, S.~Ren, and J.~Sun.
\newblock Deep residual learning for image recognition.
\newblock In {\em Proceedings of the IEEE conference on computer vision and
  pattern recognition}, pages 770--778, 2016.

\bibitem{hertzmann2000illustrating}
A.~Hertzmann and D.~Zorin.
\newblock Illustrating smooth surfaces.
\newblock In {\em Proceedings of the 27th Annual Conference on Computer
  Graphics and Interactive Techniques}, pages 517--526. ACM
  Press/Addison-Wesley Publishing Co., 2000.

\bibitem{hoiem2007recovering}
D.~Hoiem, A.~A. Efros, and M.~Hebert.
\newblock Recovering surface layout from an image.
\newblock {\em International Journal of Computer Vision}, 75(1):151--172, 2007.

\bibitem{huang20173dlite}
J.~Huang, A.~Dai, L.~J. Guibas, and M.~Nie{\ss}ner.
\newblock 3dlite: towards commodity 3d scanning for content creation.
\newblock {\em ACM Trans. Graph.}, 36(6):203--1, 2017.

\bibitem{huang2018texturenet}
J.~Huang, H.~Zhang, L.~Yi, T.~Funkhouser, M.~Nie{\ss}ner, and L.~Guibas.
\newblock Texturenet: Consistent local parametrizations for learning from
  high-resolution signals on meshes.
\newblock {\em arXiv preprint arXiv:1812.00020}, 2018.

\bibitem{huang2018quadriflow}
J.~Huang, Y.~Zhou, M.~Nie{\ss}ner, J.~R. Shewchuk, and L.~J. Guibas.
\newblock Quadriflow: A scalable and robust method for quadrangulation.
\newblock In {\em Computer Graphics Forum}, volume~37, pages 147--160. Wiley
  Online Library, 2018.

\bibitem{kahler2015very}
O.~K{\"a}hler, V.~A. Prisacariu, C.~Y. Ren, X.~Sun, P.~Torr, and D.~Murray.
\newblock Very high frame rate volumetric integration of depth images on mobile
  devices.
\newblock {\em IEEE transactions on visualization and computer graphics},
  21(11):1241--1250, 2015.

\bibitem{lai2010metric}
Y.-K. Lai, M.~Jin, X.~Xie, Y.~He, J.~Palacios, E.~Zhang, S.-M. Hu, and X.~Gu.
\newblock Metric-driven {RoSy} field design and remeshing.
\newblock {\em IEEE Transactions on Visualization and Computer Graphics},
  16(1):95--108, 2010.

\bibitem{leutenegger2011brisk}
S.~Leutenegger, M.~Chli, and R.~Siegwart.
\newblock Brisk: Binary robust invariant scalable keypoints.
\newblock In {\em 2011 IEEE international conference on computer vision
  (ICCV)}, pages 2548--2555. Ieee, 2011.

\bibitem{li2017two}
J.~Li, R.~Klein, and A.~Yao.
\newblock A two-streamed network for estimating fine-scaled depth maps from
  single rgb images.
\newblock In {\em Proceedings of the IEEE International Conference on Computer
  Vision}, pages 3372--3380, 2017.

\bibitem{lowe2004distinctive}
D.~G. Lowe.
\newblock Distinctive image features from scale-invariant keypoints.
\newblock {\em International journal of computer vision}, 60(2):91--110, 2004.

\bibitem{masci2015geodesic}
J.~Masci, D.~Boscaini, M.~Bronstein, and P.~Vandergheynst.
\newblock Geodesic convolutional neural networks on riemannian manifolds.
\newblock In {\em Proceedings of the IEEE international conference on computer
  vision workshops}, pages 37--45, 2015.

\bibitem{qi2018geonet}
X.~Qi, R.~Liao, Z.~Liu, R.~Urtasun, and J.~Jia.
\newblock Geonet: Geometric neural network for joint depth and surface normal
  estimation.
\newblock In {\em Proceedings of the IEEE Conference on Computer Vision and
  Pattern Recognition}, pages 283--291, 2018.

\bibitem{ray2009geometry}
N.~Ray, B.~Vallet, L.~Alonso, and B.~Levy.
\newblock Geometry-aware direction field processing.
\newblock {\em ACM Transactions on Graphics (TOG)}, 29(1):1, 2009.

\bibitem{ray2008n}
N.~Ray, B.~Vallet, W.~C. Li, and B.~L{\'e}vy.
\newblock $n$-symmetry direction field design.
\newblock {\em ACM Transactions on Graphics (TOG)}, 27(2):10, 2008.

\bibitem{ronneberger2015u}
O.~Ronneberger, P.~Fischer, and T.~Brox.
\newblock U-net: Convolutional networks for biomedical image segmentation.
\newblock In {\em International Conference on Medical image computing and
  computer-assisted intervention}, pages 234--241. Springer, 2015.

\bibitem{rublee2011orb}
E.~Rublee, V.~Rabaud, K.~Konolige, and G.~Bradski.
\newblock Orb: An efficient alternative to sift or surf.
\newblock 2011.

\bibitem{saxena2006learning}
A.~Saxena, S.~H. Chung, and A.~Y. Ng.
\newblock Learning depth from single monocular images.
\newblock In {\em Advances in neural information processing systems}, pages
  1161--1168, 2006.

\bibitem{shelhamer2015scene}
E.~Shelhamer, J.~T. Barron, and T.~Darrell.
\newblock Scene intrinsics and depth from a single image.
\newblock In {\em Proceedings of the IEEE International Conference on Computer
  Vision Workshops}, pages 37--44, 2015.

\bibitem{shi2015break}
J.~Shi, X.~Tao, L.~Xu, and J.~Jia.
\newblock Break ames room illusion: depth from general single images.
\newblock {\em ACM Transactions on Graphics (TOG)}, 34(6):225, 2015.

\bibitem{simo2015discriminative}
E.~Simo-Serra, E.~Trulls, L.~Ferraz, I.~Kokkinos, P.~Fua, and F.~Moreno-Noguer.
\newblock Discriminative learning of deep convolutional feature point
  descriptors.
\newblock In {\em Proceedings of the IEEE International Conference on Computer
  Vision}, pages 118--126, 2015.

\bibitem{simonyan2014very}
K.~Simonyan and A.~Zisserman.
\newblock Very deep convolutional networks for large-scale image recognition.
\newblock {\em arXiv preprint arXiv:1409.1556}, 2014.

\bibitem{song2015sun}
S.~Song, S.~P. Lichtenberg, and J.~Xiao.
\newblock Sun rgb-d: A rgb-d scene understanding benchmark suite.
\newblock In {\em Proceedings of the IEEE conference on computer vision and
  pattern recognition}, pages 567--576, 2015.

\bibitem{tatarchenko2018tangent}
M.~Tatarchenko, J.~Park, V.~Koltun, and Q.-Y. Zhou.
\newblock Tangent convolutions for dense prediction in 3d.
\newblock In {\em Proceedings of the IEEE Conference on Computer Vision and
  Pattern Recognition}, pages 3887--3896, 2018.

\bibitem{tola2010daisy}
E.~Tola, V.~Lepetit, and P.~Fua.
\newblock Daisy: An efficient dense descriptor applied to wide-baseline stereo.
\newblock {\em IEEE transactions on pattern analysis and machine intelligence},
  32(5):815--830, 2010.

\bibitem{torralba2002depth}
A.~Torralba and A.~Oliva.
\newblock Depth estimation from image structure.
\newblock {\em IEEE Transactions on pattern analysis and machine intelligence},
  24(9):1226--1238, 2002.

\bibitem{wang2019normalized}
H.~Wang, S.~Sridhar, J.~Huang, J.~Valentin, S.~Song, and L.~J. Guibas.
\newblock Normalized object coordinate space for category-level 6d object pose
  and size estimation.
\newblock {\em arXiv preprint arXiv:1901.02970}, 2019.

\bibitem{wang2016surge}
P.~Wang, X.~Shen, B.~Russell, S.~Cohen, B.~Price, and A.~L. Yuille.
\newblock Surge: Surface regularized geometry estimation from a single image.
\newblock In {\em Advances in Neural Information Processing Systems}, pages
  172--180, 2016.

\bibitem{wang2018adaptive}
P.-S. Wang, C.-Y. Sun, Y.~Liu, and X.~Tong.
\newblock Adaptive o-cnn: a patch-based deep representation of 3d shapes.
\newblock In {\em SIGGRAPH Asia 2018 Technical Papers}, page 217. ACM, 2018.

\bibitem{wang2015designing}
X.~Wang, D.~Fouhey, and A.~Gupta.
\newblock Designing deep networks for surface normal estimation.
\newblock In {\em Proceedings of the IEEE Conference on Computer Vision and
  Pattern Recognition}, pages 539--547, 2015.

\bibitem{xie2016deep3d}
J.~Xie, R.~Girshick, and A.~Farhadi.
\newblock Deep3d: Fully automatic 2d-to-3d video conversion with deep
  convolutional neural networks.
\newblock In {\em European Conference on Computer Vision}, pages 842--857.
  Springer, 2016.

\bibitem{xu2017multi}
D.~Xu, E.~Ricci, W.~Ouyang, X.~Wang, and N.~Sebe.
\newblock Multi-scale continuous crfs as sequential deep networks for monocular
  depth estimation.
\newblock In {\em Proceedings of the IEEE Conference on Computer Vision and
  Pattern Recognition}, pages 5354--5362, 2017.

\bibitem{xu2017directionally}
H.~Xu, M.~Dong, and Z.~Zhong.
\newblock Directionally convolutional networks for 3d shape segmentation.
\newblock In {\em Proceedings of the IEEE International Conference on Computer
  Vision}, pages 2698--2707, 2017.

\bibitem{yi2016lift}
K.~M. Yi, E.~Trulls, V.~Lepetit, and P.~Fua.
\newblock Lift: Learned invariant feature transform.
\newblock In {\em European Conference on Computer Vision}, pages 467--483.
  Springer, 2016.

\bibitem{yu2011asift}
G.~Yu and J.-M. Morel.
\newblock Asift: An algorithm for fully affine invariant comparison.
\newblock {\em Image Processing On Line}, 1:11--38, 2011.

\bibitem{zagoruyko2015learning}
S.~Zagoruyko and N.~Komodakis.
\newblock Learning to compare image patches via convolutional neural networks.
\newblock In {\em Proceedings of the IEEE conference on computer vision and
  pattern recognition}, pages 4353--4361, 2015.

\bibitem{zhang2018deep}
Y.~Zhang and T.~Funkhouser.
\newblock Deep depth completion of a single rgb-d image.
\newblock In {\em Proceedings of the IEEE Conference on Computer Vision and
  Pattern Recognition}, pages 175--185, 2018.

\end{thebibliography}
}
\clearpage

\section*{Supplemental}
\begin{appendix}
\section{Implementation Details}

Our best-performing implementation is based on the DORN architecture~\cite{fu2018deep}, which was originally proposed for depth estimation on the NYUv2 dataset.
We adopt the original parameters, except that we output 13 dimensions at the final layer and use our loss to train the network. During our experiments, the input and output image resolution is 320x240. We use the fixed learning rate as 1e-5.

\section{Timing}
We render one keyframe for every ten frames for ScanNet~\cite{dai2017scannet} dataset, which yields about 200,000 training samples. We take 4.8 hours to train DORN~\cite{fu2018deep} and our loss for one epoch and report the performance after training 9 epochs. During testing, we use 0.0375 second to process each image.

\section{Surface Normal Estimation}

\paragraph{Comparison with the state-of-the-art}  We show results for surface normal estimation using ScanNet for training and testing in section 5.1 of the main paper.   In this section, we do the same using the SunCG dataset~\cite{song2015sun}. Specifically, we use our approach to train four networks using the SUNCG training set and evaluate them on the SUNCG test set. Table~\ref{tab:scannet-comparison} shows the results for UNet~\cite{ronneberger2015u}, SkipNet~\cite{bansal2016marr}, GeoNet~\cite{qi2018geonet} and DORN~\cite{fu2018deep}. With the assistance of the projected tangent principal directions, the normal predictions is improved in every case.

\begin{table}[h]
    \centering
    \tabcolsep=0.08cm
    \begin{tabular}{|c|c|c|c||c|c|c|}
         \hline
         \textbf{SunCG} & mean & median & rmse & $11.25^\circ$ & $22.5^\circ$ & $30^\circ$\\
         \hline
         UNet & 14.88 & 6.20 & 24.94 & 64.4 & 78.6 & 83.9\\
         \hline
         UNet-Ours & 13.25 & 4.64 & 23.73 & 69.8 & 81.6 & 86.1\\
         \hline
         SkipNet & 13.38 & 3.97 & 24.54 & 70.2 & 80.3 & 85.1\\
         \hline
         SkipNet-Ours & 12.82 & 3.87 & 23.69 & 71.0 & 80.2 & 86.1\\
         \hline
         GeoNet & 13.14 & 3.56 & 23.54 & 70.6 & 80.7 & 86.0\\
         \hline
         GeoNet-Ours & 12.68 & 3.60 & \textbf{22.73} & 71.2 & 81.3 & \textbf{86.6}\\
         \hline
         DORN & 12.90 & 3.36 & 24.12 & 71.3 & 81.3 & 85.3\\
         \hline
         DORN-Ours & \textbf{12.38} & \textbf{3.33} & 23.34 & \textbf{72.3} & \textbf{82.3} & 86.3\\
         \hline         
    \end{tabular}
    \caption{Evaluation on Surface Normal Predictions. We train and test our algorithm with different network architectures on the SunCG~\cite{dai2017scannet}. Assisted by our joint loss, the performances of all networks are improved.}
    \label{tab:scannet-comparison}
\end{table}

\begin{figure}
    \centering
    \includegraphics[width=\linewidth]{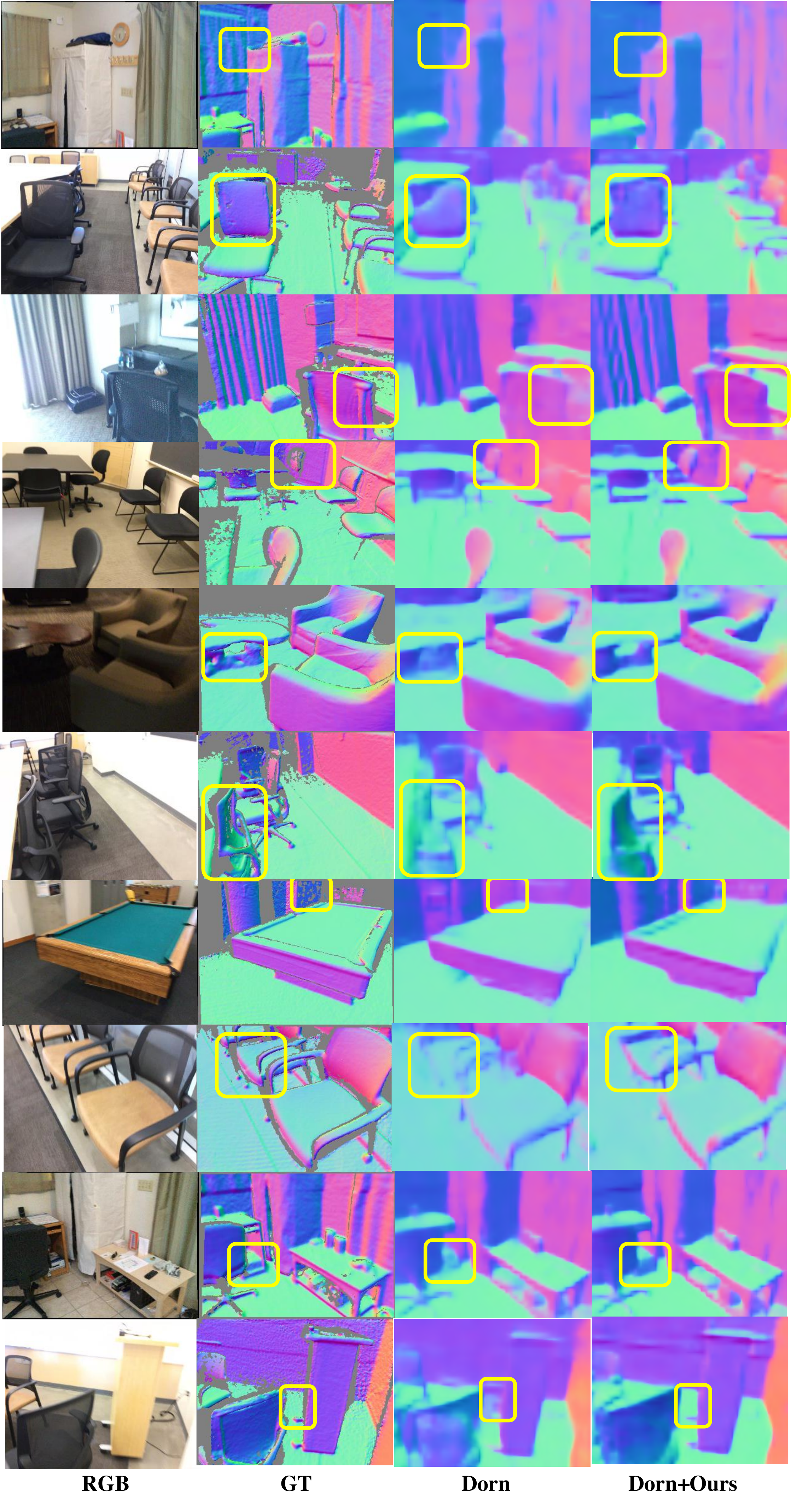}
    \caption{Visual comparison of the results. With our joint loss, the predicted surface normals produce less errors and more details. We show more accurate prediction especially for small objects.}
    \label{fig:vis-result}
\end{figure}
\begin{figure}
    \centering
    \includegraphics[width=\linewidth]{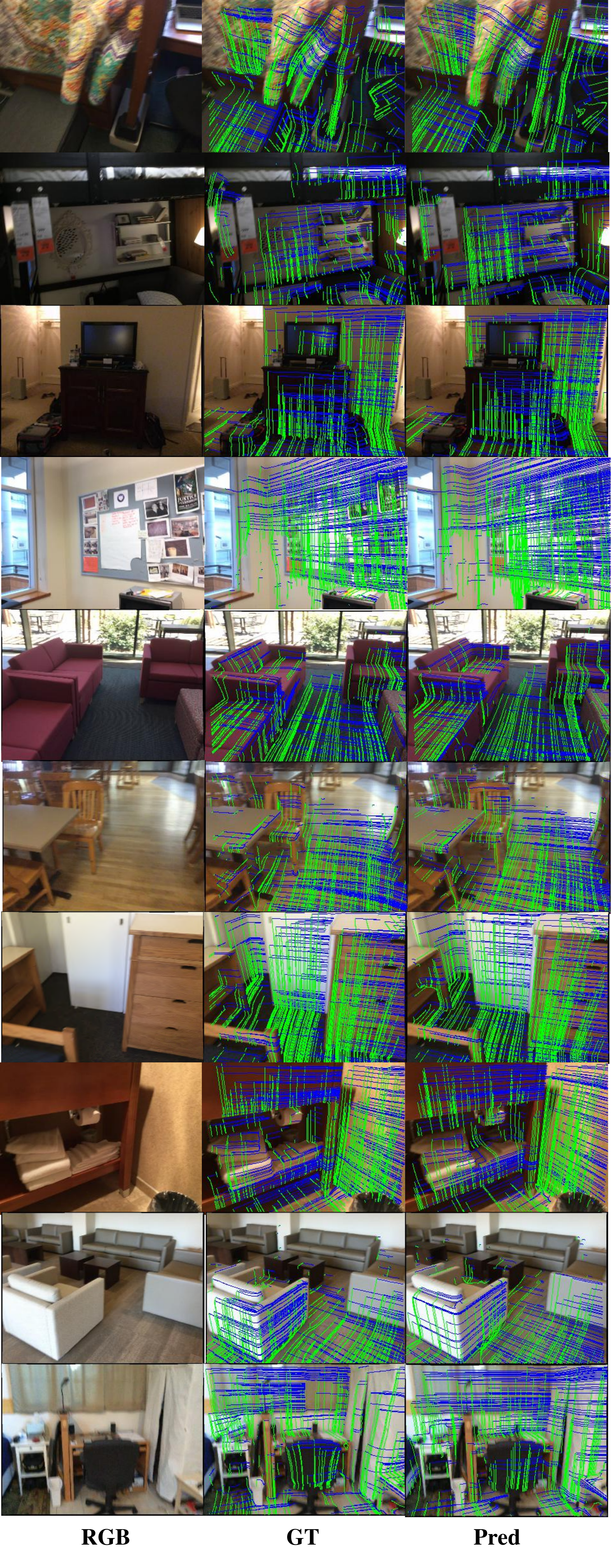}
    \caption{Visualization of the projected tangent principal directions. The visualization shows similar direction field compared to the ground truth, and is consistent with human intuition.}
    \label{fig:vis-supplemental}
\end{figure}
\begin{figure}
    \centering
    \includegraphics[width=\linewidth]{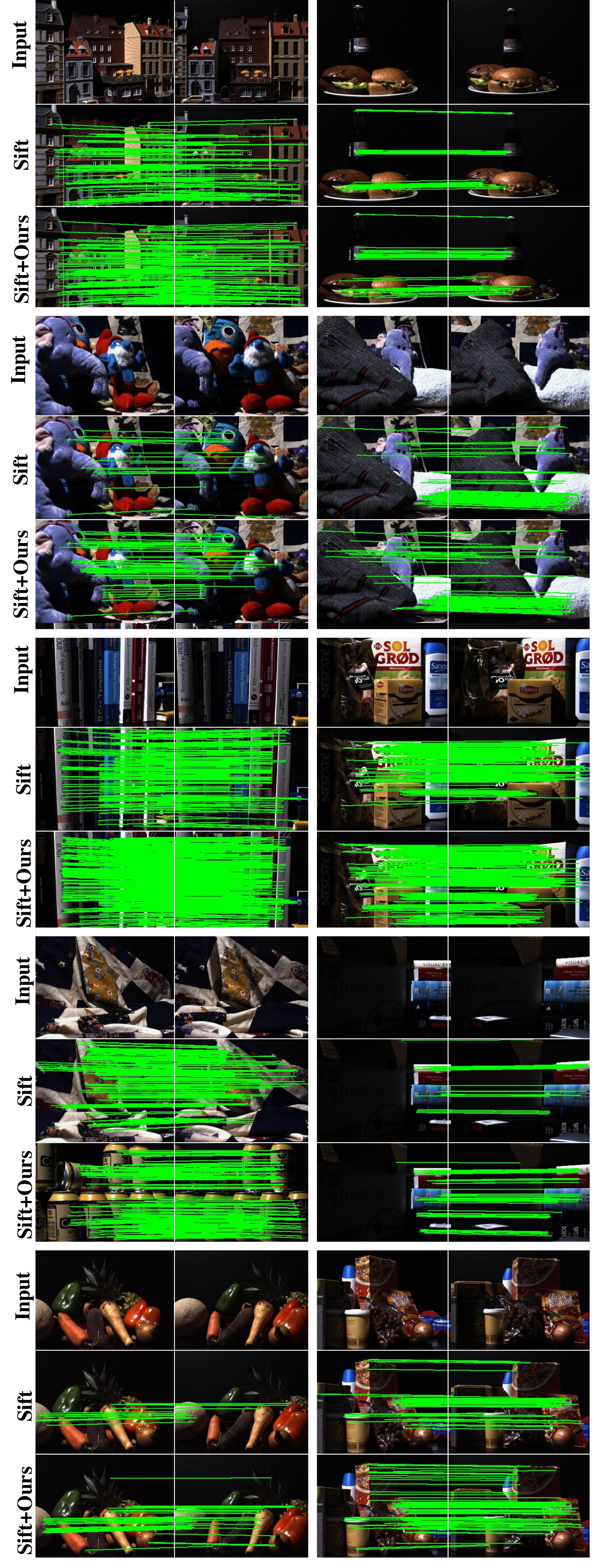}
    \caption{Visualization of the feature matching using SIFT and SIFT with our perspective rectification. We produce more correct matching than SIFT does.}
    \label{fig:dtu-supplemental}
\end{figure}
\begin{figure}
    \centering
    \includegraphics[width=\linewidth]{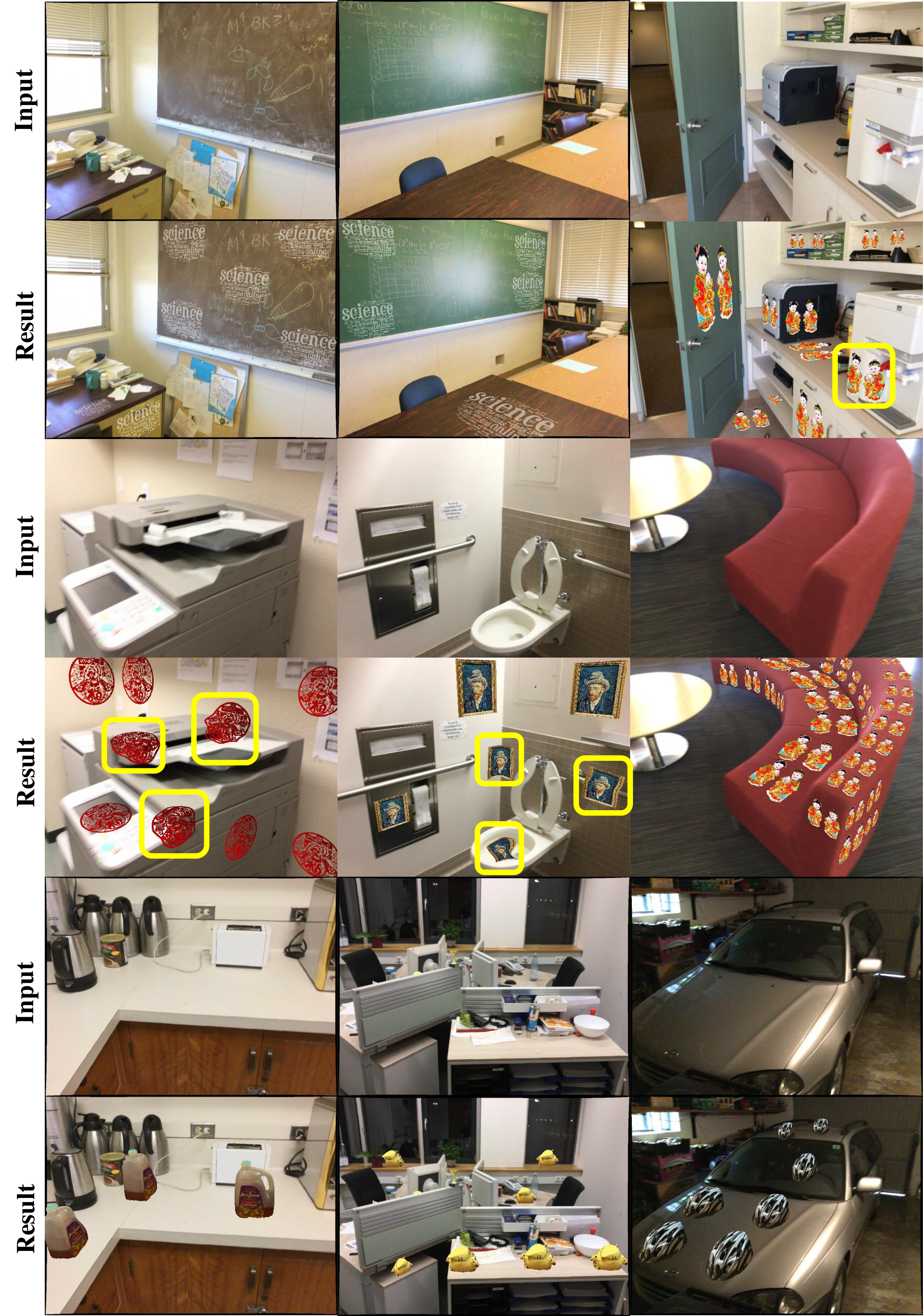}
    \caption{Visualization of augmented reality results. We attach images in a rigid or deformable way (highlighted with yellow square), or 3D objects into the scenes. The perspectives are locally consistent with the canonical frames.}
    \label{fig:ar}
\end{figure}

\section{Visualization}

\paragraph{Visualization of surface normals.} Figure~\ref{fig:vis-result} visualizes surface normal predictions using the best model with and without our approach. With our approach, the errors are smaller especially at object boundaries, possibly because of the additional supervision given by the projected tangent principal directions.  Overall, there is a noticable improvement, especially for small objects.

\paragraph{Visualization of tangent principal directions.}  Figure~\ref{fig:vis-supplemental} shows more visualizations of projected tangent principal directions predicted with our model trained on ScanNet~\cite{dai2017scannet} using  Dorn~\cite{fu2018deep} withour joint loss.  You can see that the direction fields are qualitatively similar to the ground truth and aligned with the principal directions of the surfaces.

\paragraph{Visualization of feature matching.} Figure~\ref{fig:dtu-supplemental} show more visualizations with comparisons between SIFT and SIFT with our perspective rectification on the DTU dataset~\cite{aanaes2012interesting}.  You can see that our method produces more correct matches than the baseline SIFT method.

\paragraph{Visualization of augmented reality results.} Figure~\ref{fig:ar} shows more examples of new elements inserted into images using the tangent principal directions predicted with our model. You can see that the perspectives are locally consistent with the canonical frames of the geometry.

\end{appendix}
\end{document}


\title{Supplemental Material for ``FrameNet: Learning Local Canonical Frames of 3D Surfaces
from a Single RGB Image''}

\author{First Author\\
Institution1\\
Institution1 address\\
{\tt\small firstauthor@i1.org}
\and
Second Author\\
Institution2\\
First line of institution2 address\\
{\tt\small secondauthor@i2.org}
}

\maketitle

\section{Implementation Details}
We adopt the original parameters used in DORN~\cite{fu2018deep} for NYUv2 depth estimation, except that we output 13 dimensions at the final layer and use our loss to train the network. During our experiments, the input and output image resolution is 320x240. We use the fixed learning rate as 1e-5.

\section{Surface Normal Estimation}
\paragraph{Compare with the state-of-the-art} We compare the performance of the surface normal estimation from our approach with the state-of-the-art methods on SunCG~\cite{song2015sun}. We use our approach to train four networks and evaluate them. Table~\ref{tab:scannet-comparison} shows the results including UNet~\cite{ronneberger2015u}, SkipNet~\cite{bansal2016marr}, GeoNet~\cite{qi2018geonet} and DORN~\cite{fu2018deep}. With the assistance of the projected tangent principal directions, the normal prediction has been improved. Please refer to the experiments for ScanNet in the main paper in section 5.1.
\begin{table}
    \centering
    \tabcolsep=0.08cm
    \begin{tabular}{|c|c|c|c||c|c|c|}
         \hline
         \textbf{SunCG} & mean & median & rmse & $11.25^\circ$ & $22.5^\circ$ & $30^\circ$\\
         \hline
         UNet & 14.88 & 6.20 & 24.94 & 64.4 & 78.6 & 83.9\\
         \hline
         UNet-Ours & 13.25 & 4.64 & 23.73 & 69.8 & 81.6 & 86.1\\
         \hline
         SkipNet & 13.38 & 3.97 & 24.54 & 70.2 & 80.3 & 85.1\\
         \hline
         SkipNet-Ours & 12.82 & 3.87 & 23.69 & 71.0 & 80.2 & 86.1\\
         \hline
         GeoNet & 13.14 & 3.56 & 23.54 & 70.6 & 80.7 & 86.0\\
         \hline
         GeoNet-Ours & 12.68 & 3.60 & \textbf{22.73} & 71.2 & 81.3 & \textbf{86.6}\\
         \hline
         DORN & 12.90 & 3.36 & 24.12 & 71.3 & 81.3 & 85.3\\
         \hline
         DORN-Ours & \textbf{12.38} & \textbf{3.33} & 23.34 & \textbf{72.3} & \textbf{82.3} & 86.3\\
         \hline         
    \end{tabular}
    \caption{Evaluation on Surface Normal Predictions. We train and test our algorithm with different network architectures on the SunCG~\cite{dai2017scannet}. Assisted by our joint loss, the performances of all networks are improved.}
    \label{tab:scannet-comparison}
\end{table}

\paragraph{Test on NYUv2} We test different versions of our network on NYUv2~\cite{eigen2014depth} as a standard evaluation dataset. We train the network on SunCG datasets and directly test on NYUv2, as shown in Table~\ref{tab:vis-nyu}. Specifically, GeoNet-origin trained and tested on NYUv2~\cite{qi2018geonet}, and is the current state-of-the-art method on normal estimation. Other rows are networks trained w/o. our joint losses on SunCG.

\begin{table}
    \centering
    \tabcolsep=0.07cm
    \begin{tabular}{|c|c|c|c||c|c|c|}
        \hline
         \textbf{NYUv2} & mean & median & rmse & $11.25^\circ$ & $22.5^\circ$ & $30^\circ$\\
         \hline
         GeoNet-origin & \textbf{19.0} & \textbf{11.8} & \textbf{26.9} & \textbf{48.4} & \textbf{71.5} & \textbf{79.5}\\
         \hline
         \hline
         \textbf{SunCG} & mean & median & rmse & $11.25^\circ$ & $22.5^\circ$ & $30^\circ$\\
         \hline
         UNet & 25.21 & 18.26 & 32.82 & 32.2 & 57.7 & 68.3\\
         \hline
         UNet-Ours & 24.64 & 17.10 & 32.65 & 35.0 & 59.6 & 69.5\\
         \hline
         SkipNet & 24.75 & 17.36 & 32.45 & 33.8 & 58.1 & 69.0\\
         \hline
         SkipNet-Ours & 23.67 & 16.28 & 31.72 & 36.1 & 62.2 & 72.7\\
         \hline
         GeoNet & 22.32 & 14.97 & 30.59 & 39.8 & 64.3 & 73.4\\
         \hline
         GeoNet-Ours & 22.15 & 14.41 & 30.18 & 40.1 & 65.3 & 74.4\\
         \hline
         DORN & 22.19 & 14.46 & 30.16 & 40.3 & 65.3 & 74.1\\
         \hline
         DORN-Ours & 21.99 & 14.29 & 29.87 & 40.5 & 65.8 & 74.6\\
         \hline         
    \end{tabular}
    \caption{Normal prediction on NYUv2~\cite{eigen2014depth}. GeoNet-origin trained and tested on NYUv2~\cite{qi2018geonet}. In other rows, we train network w/o. our joint loss on SunCG and tested on NYUv2. DORN-Ours trained on ScanNet performs best among all.}
    \label{tab:vis-nyu}
\end{table}

The joint loss in the training process results in better normal estimation. From the ScanNet experiment in section 5.1 of the main section,  ScanNet gets better performance compared to SunCG, possibly due to the domain gap between synthetic (SunCG) and real (NYUv2).

\begin{figure}
    \centering
    \includegraphics[width=\linewidth]{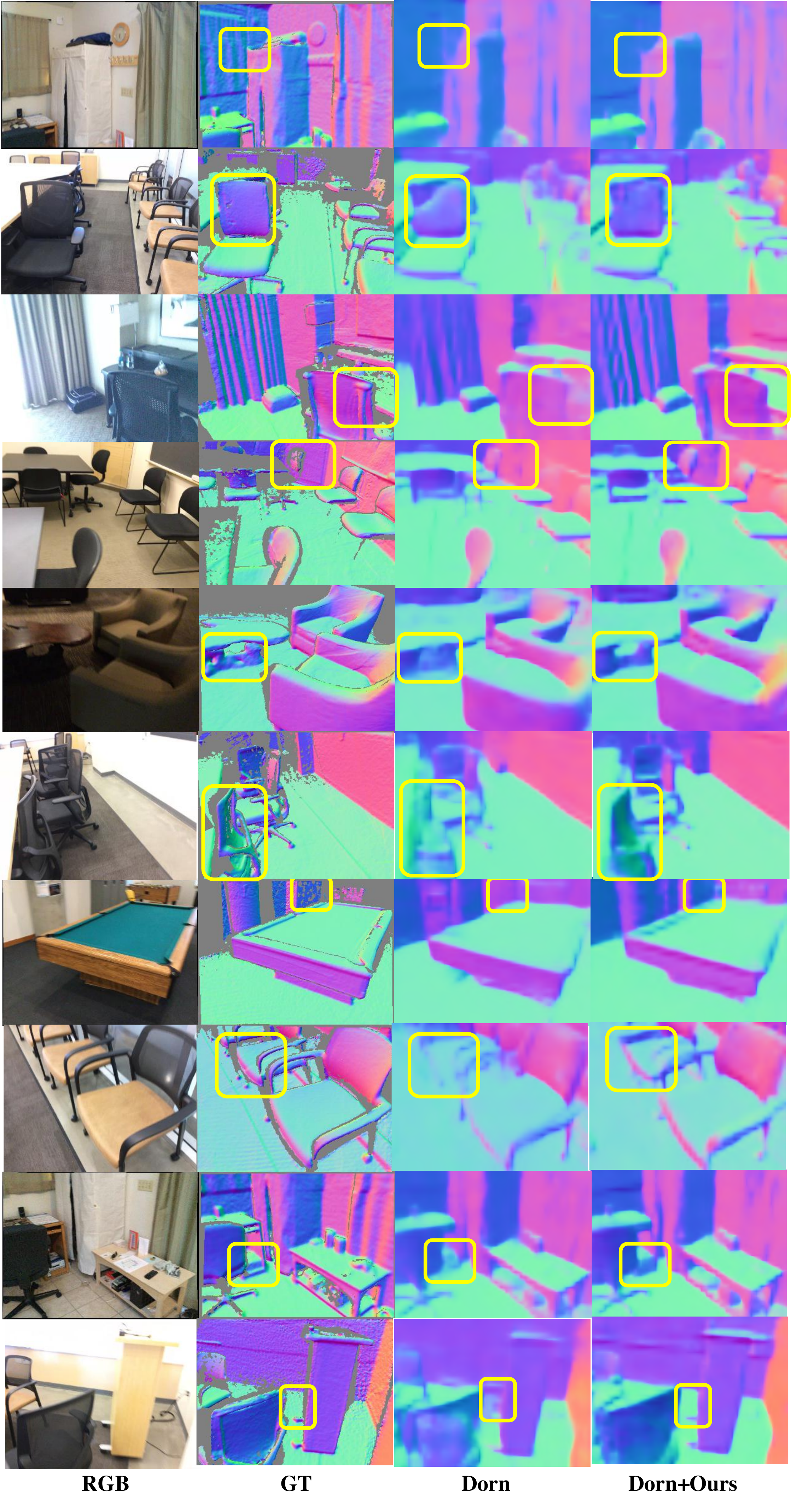}
    \caption{Visual comparison of the results. With our joint loss, the predicted surface normals produce less errors and more details. We show more accurate prediction especially for small objects.}
    \label{fig:vis-result}
\end{figure}
\begin{figure}
    \centering
    \includegraphics[width=\linewidth]{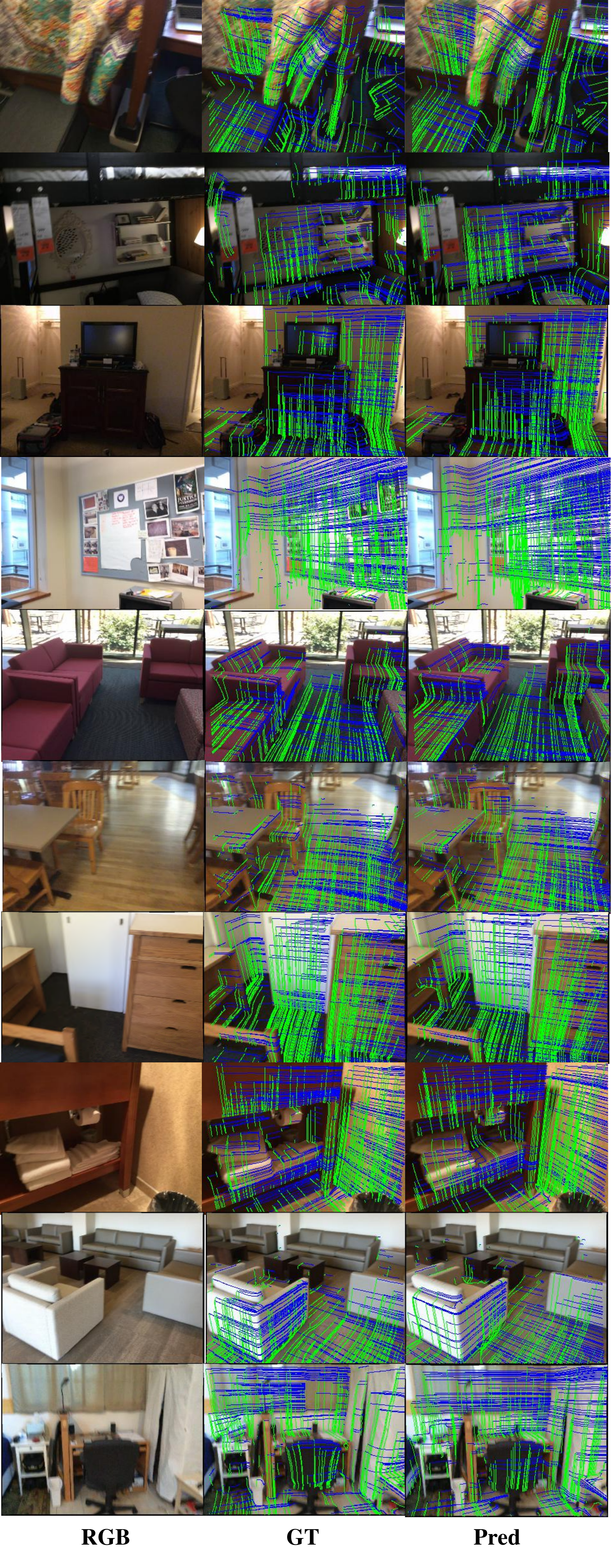}
    \caption{Visualization of the projected tangent principal directions. The visualization shows similar direction field compared to the ground truth, and is consistent with human intuition.}
    \label{fig:vis-supplemental}
\end{figure}
\begin{figure}
    \centering
    \includegraphics[width=\linewidth]{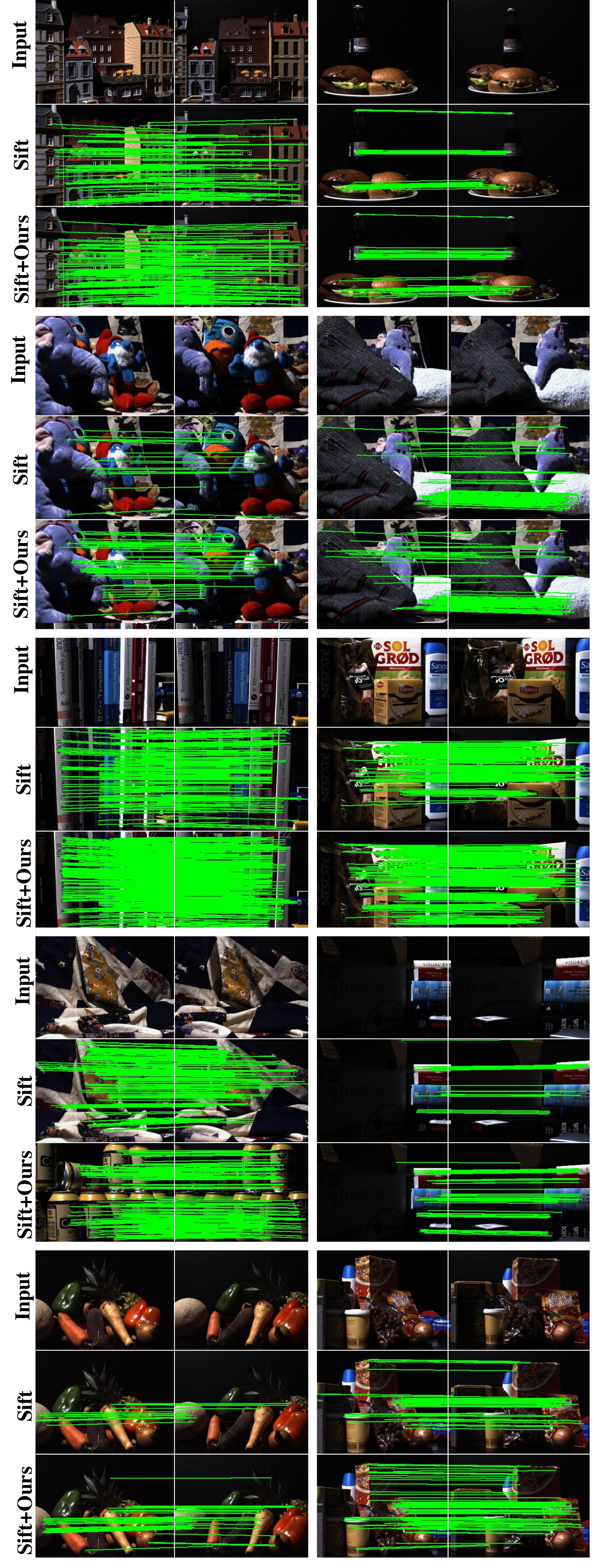}
    \caption{Visualization of the feature matching using SIFT and SIFT with our perspective rectification. We produce more correct matching than SIFT does.}
    \label{fig:dtu-supplemental}
\end{figure}
\begin{figure}
    \centering
    \includegraphics[width=\linewidth]{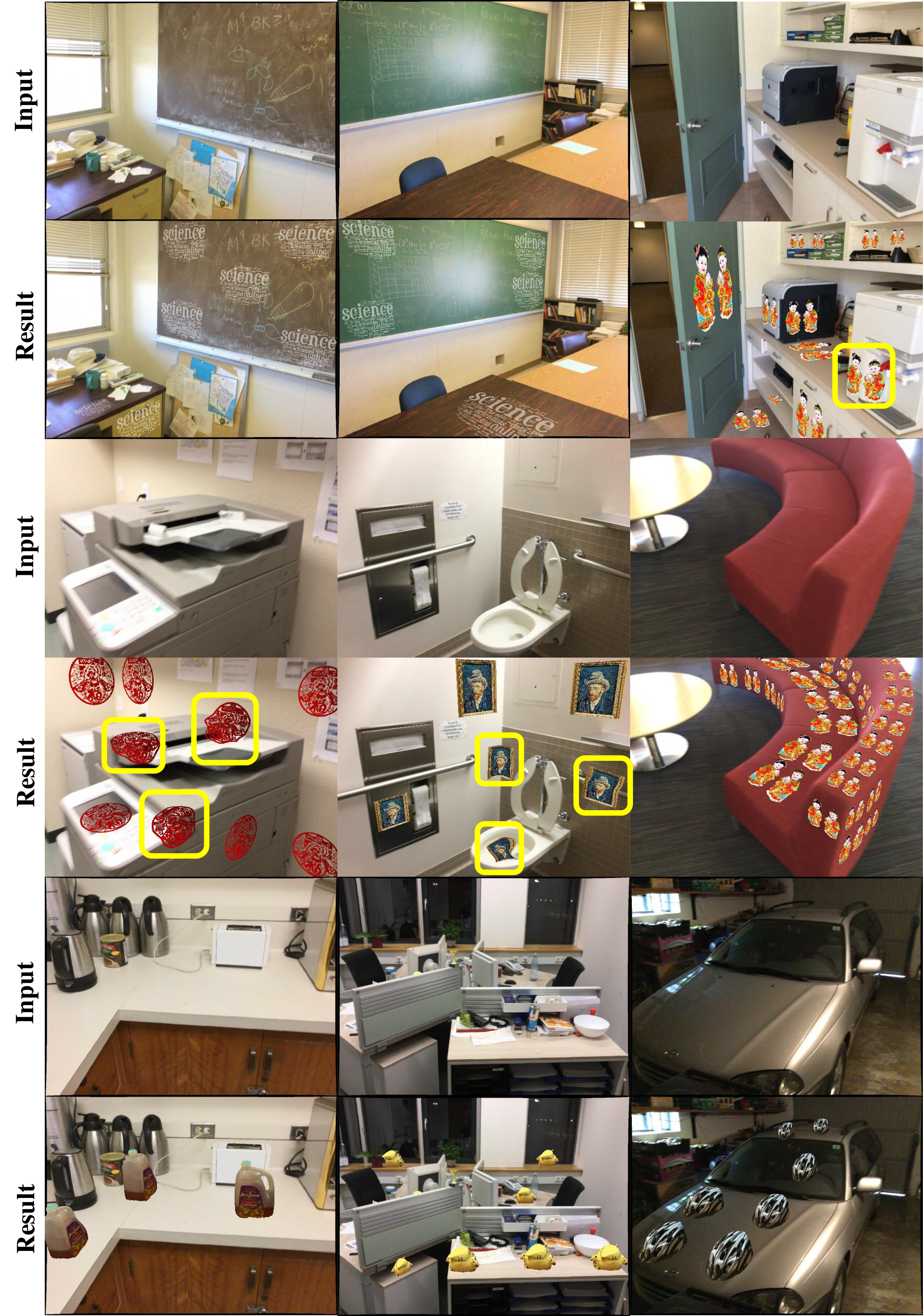}
    \caption{Visualization of augmented reality results. We attach images in a rigid or deformable way (highlighted with yellow square), or 3D objects into the scenes. The perspectives are locally consistent with the canonical frames.}
    \label{fig:ar}
\end{figure}

\section{Visualization}
\paragraph{Surface Normal Comparison} Figure~\ref{fig:vis-result} visualizes the normal prediction using the best model w/o. our approach on both the datasets. With our approach, the errors are smaller especially at object boundaries, possibly because of the additional supervision given by the projected tangent principal directions. We show more accurate prediction, especially for small objects.

\paragraph{Visualize the tangent principal directions} We show more visualization for the projected tangent principal directions in figure~\ref{fig:vis-supplemental}. The model is trained using the Dorn~\cite{fu2018deep} with our joint loss on ScanNet~\cite{dai2017scannet}. The visualization shows a similar direction field compared to the ground truth and is consistent with human intuition.

\paragraph{Visualize the feature matching} We show more visualization for comparison between SIFT and SIFT with our perspective rectification on the DTU~\cite{aanaes2012interesting} in figure~\ref{fig:dtu-supplemental}. We produce more correct matching than SIFT does.

\paragraph{Visualize the augmented reality results} We show more examples of new elements insertion into the scene in figure~\ref{fig:ar}. The perspectives are locally consistent with the canonical frames of the geometry.
\clearpage
{\small
\bibliographystyle{ieee}
\bibliography{egbib}
}